# Conditions for energetically-optimal elasticity and their implications for biomimetic propulsion systems

Arion Pons and Tsevi Beatus

**Abstract:** Minimising the energy consumption associated with periodic motion is a priority common to a wide range of technologies and organisms – among them, many species of flying insect, for which flapping-wing flight is a life-essential mode of locomotion. In pursuit of this priority, the following problem often manifests: how to introduce elasticity into an actuated, oscillating, system in order to minimise actuator power consumption? Here, we explore this question in a range of general systems, and find some surprising answers. For instance, it is widely assumed that, if the system dynamics are linear, then linear resonant elasticity is the only optimal choice. We show, to the contrary, that there exist nonlinear elasticities with equivalent optimality, and provide an elegant method for constructing these elasticities in general systems. This is a new principle of linear and nonlinear dynamics, fundamentally altering how questions of energetic optimality in a wide range of dynamical systems must be approached. Furthermore, we show how this principle enables new forms of optimal system design, including optimal unidirectional actuation in nonlinear systems; new tools for the design of optimal biomimetic propulsion systems; and new insights into the role of structural elasticity in a range of different organisms.

## 1. Introduction

Periodic motion is ubiquitous, in both natural and manufactured systems. In many cases, this motion is intentional – *i.e.*, driven by an actuator, and powered by a source of energy – and in such cases, the question of how to minimise the consumption of this energy often manifests. In studies of biological and biomimetic systems for propulsion within a fluid, for instance, the question of how energy consumption is minimised is highly relevant: in jet-propelled jellyfish and squid [1–5]; thunniform and carangiform aquatic animals [6, 7]; microorganisms with flagellar or ciliary motors [8, 9]; flapping-wing insects [10–17] or biomimetic hummingbirds [18]; and further. Indeed, these studies are united by a more specific commonality: in each, energetic optimality is connected with the phenomenon of resonance; referring to the way in

A. Pons (✉) · T. Beatus (✉)
The Silberman Institute of Life Sciences, Hebrew University of Jerusalem, Giv'at Ram, Jerusalem, Israel.
email: arion.pons@mail.huji.ac.il

A. Pons · T. Beatus
The Benin School of Computer Science and Engineering; Hebrew University of Jerusalem, Giv'at Ram, Jerusalem, Israel.

T. Beatus
tsevi.beatus@mail.huji.ac.il

which structural elasticity, linear or nonlinear, can absorb inertial loads, reduce system power requirements, and/or maximise the system's response to input forces.

The ways in which resonant effects manifest in biological propulsion systems are varied and often uncertain: for instance, in a range of insect species, resonant effects are thought to play a core role within the flight motor system [10–17]; but the mechanisms for controlling this resonant behaviour are unclear. Complex and counterintuitive forms of control are observed in several species that are thought to utilise resonant effects. For body pitch-angle control, independent modulation of only the forward-stroke position – *i.e.*, simultaneous change in stroke amplitude and static offset – is observed [19, 20]. For position and payload control, wingbeat frequency modulation is observed, even when amplitude modulation is available for apparently the same purposes [14, 16, 21]. Do these species ignore the anticipated energetic cost of deviating from a resonant state? [14]. Or are other control mechanisms at work – for instance, active elasticity control? [16, 22]. Similar questions may be found in a range of disparate fields: questions regarding the energetic effects, and limitations, of resonance in jellyfish propulsion [4, 5]; limb and leg joint behaviour [23–25]; energy harvesters [26–30]; and vibro-impact drilling systems [31–33].

To probe a commonality of these topics – the energetic role of elasticity in oscillating systems – we study an oscillating system of general form. This system is driven by an actuator, under the conditions of a particular given kinematic output (desired, or prescribed) and particular load requirements (inertial, dissipative, *etc.*). The option to introduce elasticity into the system is available. In this context, we consider a fundamental question. What elasticity – if any – is best to introduce, in order to optimise overall actuator power consumption? This question represents an inverse problem: a system characteristic (elasticity) is to be chosen in order to optimally replicate specified behaviour; an inversion of the normative solution order, in which behaviour is predicted from specified characteristics.

If the oscillating system in question has linear inertia and damping, and the available elasticity is restricted to linear elastic elements, then the answer to this question is well known. Linear resonance, with a linear elastic element tuned to a system resonant frequency, in series or in parallel with the actuator, is energetically optimal [34]. However, to the authors' knowledge, a more general form of this problem has not previously been considered. Suppose that, in a system with linear inertia and dissipation, any conceivable nonlinear elasticity is permitted? With increasing capability in additive manufacturing and compliant systems design – *e.g.*, for biomimetic flapping-wing micro-air vehicles (FW-MAVs) [35] – there is a little *a priori* need to restrict the system to linearity. The resonant linear elasticity remains energetically optimal – but, as per our question, do other equally optimal nonlinear elasticities exist? We provide the first answer to this question: yes, there are nonlinear elasticities of equal optimality, in terms of overall actuator power consumption. We go on to derive simple, general, and intuitive methods for constructing these optimal nonlinear elasticities – in linear, and nonlinear systems. In a wide range of systems, we find that conditions for optimality reduce to simple bounds on the space of elasticity: the elastic-bound conditions. We show how, and under what circumstances, these conditions propagate through the actuator drivetrain; from optimality in terms of mechanical power; to optimality in terms of electrical, chemical, or metabolic power. These results for elastic optimality have significant implications for the design and control of a huge range of theoretical and practical systems: not only FW-MAVs; but micro-swimmers, energy harvesters, vibro-impact drills, and a wide range of other electromechanical oscillators.

## 2. Elastic-bound conditions

### 2.1. Metrics of mechanical power

In the drivetrain of a system undergoing forced periodic motion, energy, or power, flows from some designated source (*e.g.*, an electrical cell, or biological metabolism) to some designated output (*e.g.*, system kinetic energy, heat dissipation, *etc.*): Fig. 1.A shows a schematic. Between source and output, it is common to find energy in the form of mechanical power – for instance, the mechanical power required for wingbeat motion in an insect or FW-MAV [36–38]; for leg motion in a bipedal organisms and robots [39, 40]; or for vibrational motion of a vibro-impact drill [33]. This mechanical power requirement can be quantified by a range of metrics.

For a single-degree-of-freedom (1DOF) actuator undergoing periodic motion – with actuator displacement $x(t)$, time $t$, period $T$ – the mechanical power requirement is $P(t) = F(t)\dot{x}(t)$, for actuator force $F(t)$. This power requirement, $P(t)$, may contain regions of both positive ($P > 0$) and negative ($P < 0$) power. Under our convention, positive power ($P > 0$) represents power flow from actuator to system; and negative power ($P < 0$), vice-versa. By definition, areas of positive power represent an unavoidable power requirement for the actuator; but the implications of negative power are actuator-specific [36, 41, 42]: must the actuator provide this power? can it dissipate it? can it store it? In light of these questions, four broad metrics of overall mechanical power consumption, $\overline{P}_{(\cdot)}$ can be identified:

Metric (**a**): the net power:

$$\overline{P}_{(a)} = \frac{1}{T}\int_0^T P(t)\, dt, \tag{1}$$

representing the overall rate of energy dissipation in the system. Conservative forces, *e.g.*, linear or nonlinear elasticity, have no effect on the net power. According to Eq. 1, negative power ($P < 0$) is credited directly against positive power ($P > 0$), *i.e.*, negative power is perfectly absorbed and released [36, 38]. This makes Eq. 1 an unsuitable metric for characterizing resonance in a system with prescribed kinematics, because all possible elasticities generate identical values (see appendix, A.3-A.4).

Metric (**b**): the absolute power:

$$\overline{P}_{(b)} = \frac{1}{T}\int_0^T |P(t)|\, dt. \tag{2}$$

Under this metric, the actuator must provide both positive and negative power at full energetic cost. For example, consider a spacecraft reaction engine: the positive power of spacecraft acceleration, and the negative power of spacecraft braking, must both be provided by the engine. This metric is commonly used in robotics [43, 44] and biomechanics [40, 45–47], and its minimisation is a motivation for the absorption and release of negative power. In some systems, however, negative power may incur a lesser energetic cost [42], hence:

Metric (**c**): the positive-only power:

$$\overline{P}_{(c)} = \frac{1}{T}\int_0^T P(t)[P(t) \geq 0]_{\mathbb{I}}\, dt, \tag{3}$$

where $[\,\cdot\,]_{\mathbb{I}}$ is the Iverson bracket, with $[\lambda]_{\mathbb{I}} = 1$ for a true statement $\lambda$, and $[\lambda]_{\mathbb{I}} = 0$ for a false statement $\lambda$ [48], Under this metric, the actuator must provide only positive power: it is implied that the system contains some mechanism to perfectly dissipate, but not store, negative power. For example, consider a braking car, where friction within the brakes dissipates all negative power. This metric is used in the study of insect flight energetics [36, 37, 49] and the biomechanics of bipedal walking [39].

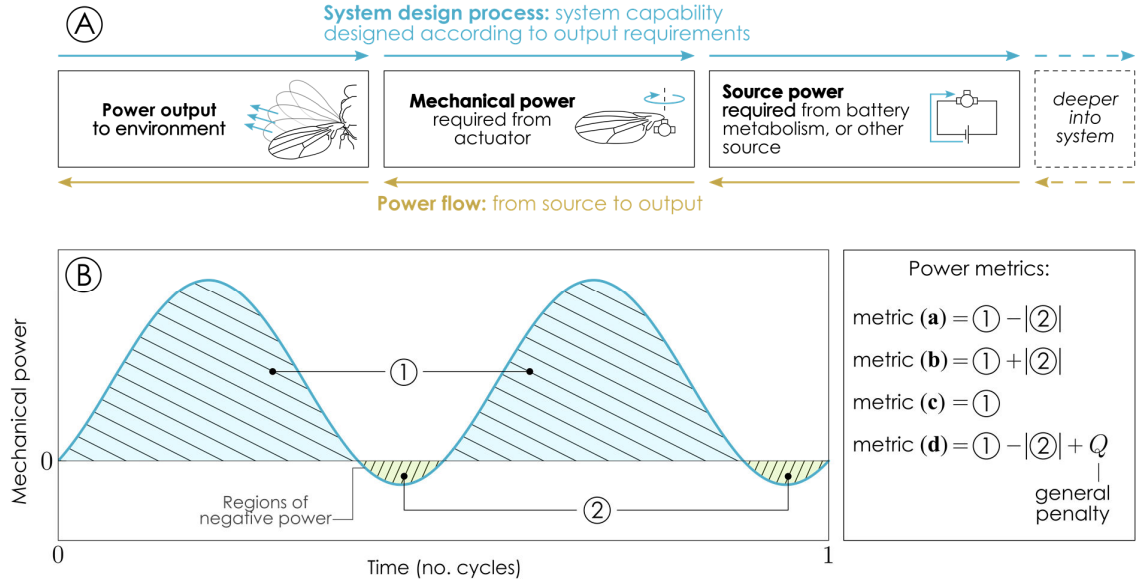

**Fig. 1. General system power behaviour.** (**A**) Schematic of system power flow: the direction of power flow, and the direction of the design process, are often opposed. (**B**) Visualisation of general periodic mechanical power requirement, $P(t)$, with overall metrics (**a**)-(**d**). Areas of positive (1) and negative (2) power are indicated, as are their effect on the four overall mechanical power metrics.

In fact, a general metric, (**d**), can be defined to encompass all three metrics, (**a**)-(**c**), and a range of other realistic actuator behaviour:

$$\overline{P}_{(d)} = \frac{1}{T}\int_0^T \left(P(t) + Q(t)|P(t)|[P(t) \leq 0]_\mathbb{I}\right) dt, \tag{4}$$

for a general penalty function $Q(t)$, penalising regions of negative power ($P < 0$). For $\overline{P}_{(d)}$ to be intelligible as a metric of actuator power consumption, $Q(t) > 0, \forall t$. Metrics (**a**)-(**c**) can be recovered with the constant penalty functions:

$$(\mathbf{a})\ Q = 0, \quad (\mathbf{b})\ Q = 2, \quad (\mathbf{c})\ Q = 1, \tag{5}$$

and other metrics, involving partial penalisation of negative power, or imperfect dissipative braking, can be defined as needed. For instance, models of metabolic cost, based on mechanical power, have been proposed, with penalties equivalent to $Q = 1.33$ and $Q = 1.20$ [50–52]. The commonality between all these metrics is that they represent differing forms of negative work penalisation: differing answers to the question of how an actuator responds to a negative power requirement. It follows – and we shall later demonstrate in detail – that in the absence of negative power ($P(t) \geq 0, \forall t$), metrics (**a**)-(**d**) are all equivalent. This is extended form of the global resonance condition devised by Ma and Zhang [27–30] to characterise vibrational energy harvesting.

## 2.2. Optimality in parallel-elastic actuation (PEA) systems

Unlike in the case of the net power, metric (**a**), the effect of elasticity on mechanical power metrics (**b**)-(**d**) is system-dependent. Consider a general nonlinear time-invariant single-degree-of-freedom (1DOF) parallel-elastic actuation (PEA) system: an actuator and an elastic element are linked to the system in parallel. As per Fig. 2, for a system with general Newtonian dynamics $D(\cdot)$, the equations of motion read:

$$\begin{aligned} G(t) &= D(x, \dot{x}, \ddot{x}, \dots), & \text{without elasticity} \\ F(t) &= D(x, \dot{x}, \ddot{x}, \dots) + F_s(x), & \text{with elasticity} \end{aligned} \quad (6)$$

where $F(t)$ and $G(t)$ are the actuator loads with and without elasticity, respectively; $x(t)$ is system displacement (or, in general, kinematic output); and $F_s(x)$ is the elastic profile, dependent only on $x$. Two points should be noted. Firstly, the dynamical formulation $D(\cdot)$ is chosen to allow general treatment of linear and nonlinear systems: for instance, as in Fig. 2, a system with linear inertia and dissipation can be represented as $D(\dot{x}, \ddot{x}) = m\ddot{x} + c\dot{x}$; and a system with linear inertia and quadratic dissipation as $D(\dot{x}, \ddot{x}) = m\ddot{x} + c\dot{x}|\dot{x}|$. Secondly, the elastic profile is formulated such that $F_s(x)$ is a nonlinear generalisation of a linear elasticity, $kx$: a linear PEA system with linear elasticity reads $m\ddot{x} + c\dot{x} + kx = F(t)$; and its analogue with nonlinear elasticity, $m\ddot{x} + c\dot{x} + F_s(x) = F(t)$.

In our analysis, the system kinematic output, $x(t)$, is prescribed. It is periodic – *i.e.*, a steady-state cycle – with period $T$, and spans the kinematic range $x(t) \in [x_1, x_2]$. In this context, $D(x, \dot{x}, \ddot{x}, \dots)$ can be computed directly, and the actuator load $G(t)$ can be visualized in load-displacement space ($G$-$x$, *i.e.*, ordinate $G$, abscissa $x$): this reveals the work, or hysteresis, loop associated with the actuator load [53, 54]. In load-displacement space, area, $dx \cdot dF$, is mechanical work: providing a helpful way to visualise and analyse the effect of elasticity on metrics of mechanical power, (**a**)-(**d**). We require only a few conditions on the work loop associated with the inelastic system load, $G(t)$: it must be a closed simple curve (*i.e.*, no self-intersection), no more than bivalued at any $x \in [x_1, x_2]$; and it must represent net power dissipation (*i.e.*, the progression of time must represent clockwise travel around the closed loop). These conditions are satisfied, *e.g.*, in any system (**i**) with positive linear inertia; (**ii**) with positive linear or nonlinear dissipation; and (**iii**) undergoing a periodic kinematic waveform, $x(t)$, composed of two half-cycles that are each monotonic[1].

Under these general conditions, we may segment the inelastic work loop, $G$, into upper ($G^+$) and lower ($G^-$) single-valued curves, associated with separate oscillatory half cycles (Fig. 2). These, in turn, generate upper ($F^+$) and lower ($F^-$) single-valued curves associated with the elastic work loop, $F$. Both pairs of curves can be defined with a midline ($G_{\text{mid}}$) and half-width ($G_{\text{arc}}$) component, such that:

$$\begin{aligned} G^{\pm}(x) &= G_{\text{mid}}(x) \pm G_{\text{arc}}(x), \\ F^{\pm}(x) &= G_{\text{mid}}(x) \pm G_{\text{arc}}(x) + F_s(x) = G^{\pm}(x) + F_s(x), \end{aligned} \quad (7)$$

over $x \in [x_1, x_2]$, and where $G_{\text{arc}}(x) \geq 0$, $\forall x$, by definition. Under a change of variables ($dx = \dot{x} \cdot dt$), the mechanical power consumption metrics (**a**)-(**d**) can then be expressed in terms of the loop parameters in Eq. 7 – *e.g.*, for $F^{\pm}$, as:

---

[1] But, not necessarily symmetric – in time, amplitude, or any other way.

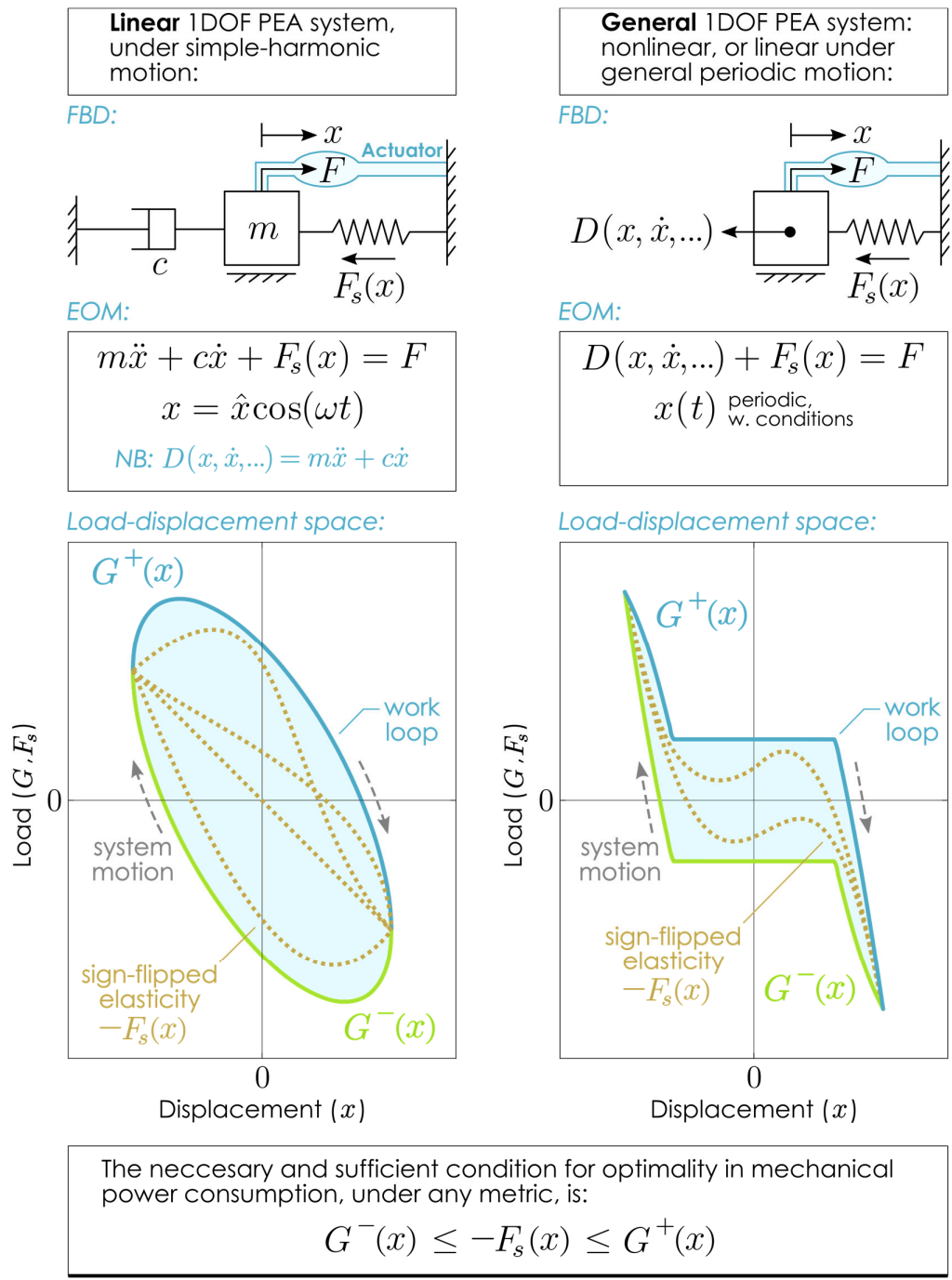

**Fig. 2. Elastic-bound conditions in linear and nonlinear PEA systems.** Free-body diagrams (FBD) and equations of motion (EOM) for linear and nonlinear PEA systems, with the elastic-bound conditions (Eq. 10) illustrated on example work loops in both cases. Note that the sign-flipped elasticity, $-F_s(x)$, can be identified with the elastic reaction load.

$$\overline{P}_{(a)} = \frac{1}{T}\int_{x_1}^{x_2} (F^+ - F^-)\,dx,$$

$$\overline{P}_{(b)} = \frac{1}{T}\int_{x_1}^{x_2} (|F^+| + |F^-|)\,dx,$$

$$\overline{P}_{(c)} = \frac{1}{T}\int_{x_1}^{x_2} (F^+[F^+ \geq 0]_\mathbb{I} - F^-[F^- \leq 0]_\mathbb{I})\,dx, \qquad (8)$$

$$\overline{P}_{(d)} = \frac{1}{T}\int_{x_1}^{x_2} (F^+ - F^- + Q^+|F^+|[F^+ \leq 0]_\mathbb{I} + Q^-|F^-|[F^- \geq 0]_\mathbb{I})\,dx,$$

where $Q^\pm(x) > 0$ are power penalty functions as per the metric **(d)**. Metrics **(a)**-**(c)** are recovered in **(d)** with:

$$\textbf{(a) } Q^\pm = 0, \quad \textbf{(b) } Q^\pm = 2, \quad \textbf{(c) } Q^\pm = 1. \qquad (9)$$

Finally, we may ask the question of elastic optimality: what are the elastic profiles, $F_s(x)$, such that mechanical power consumption, in cases **(b)**-**(d)**, is minimised? A simple answer exists:

$$\boxed{G^-(x) \leq -F_s(x) \leq G^+(x), \qquad x \in [x_1, x_2].} \qquad (10)$$

Eq. 10 is the sole criterion, necessary and sufficient, for optimality, under metrics **(b)**-**(d)**. That is, any sign-flipped nonlinear elastic load profile, $-F_s(x)$, that lies within the bounds of the inelastic work loop, $G^\pm(x)$, is optimal in mechanical power consumption, irrespective of the penalties associated with negative power. Figure 2 shows a schematic of these conditions, in linear and general nonlinear PEA systems. Eq. 10 is a powerful and general result that allows us to construct a wide range of energetically-optimal systems: a full proof, as well as a derivation of $G^\pm(x)$ in specific cases, is presented in the appendix (A.1, A.3). As an aid to understanding, however, an intuitive, geometric, explanation can be made, as illustrated in Fig. 3. This explanation may be summarised in the following three-step process:

**Step 1: The absence of negative power leads to equality among power metrics.**
As noted in Section 2.1, for any actuator load given in the time-domain, such as $G(t)$ or $F(t)$, regions of negative power are generated at times when the load, $G(t)$ or $F(t)$, and velocity, $\dot{x}(t)$, have opposite signs. In equivalent work-loop terms, regions of negative power are generated where the upper work loop curve, $G^+(x)$ or $F^+(x)$, becomes negative; or the lower curve, $G^-(x)$ or $F^-(x)$, becomes positive, as illustrated in Fig. 3A. These regions of negative power, if they exist, are associated with penalties in the power metrics **(b)**-**(d)**, as per Eqs. 8-9. For illustration, Fig. 3A shows the penalties associated with metrics **(b)** and **(c)** as equivalent areas in load-displacement space. In any given work loop, these penalties will always ensure that metrics **(b)**-**(d)** take values larger than or equal to the value of the net power, metric **(a)**. The net power is simply the area of the work loop; and these penalties can only add more area, *i.e.*, additional power requirement. However, because these penalties are dependent on the existence of negative power, if any given loop contains no regions of negative power, then, for that loop, all metrics **(a)**-**(d)** are equivalent: $\overline{P}_{(a)} = \overline{P}_{(b)} = \overline{P}_{(c)} = \overline{P}_{(d)}$; an extension of the global resonance condition of Ma and Zhang [27–30].

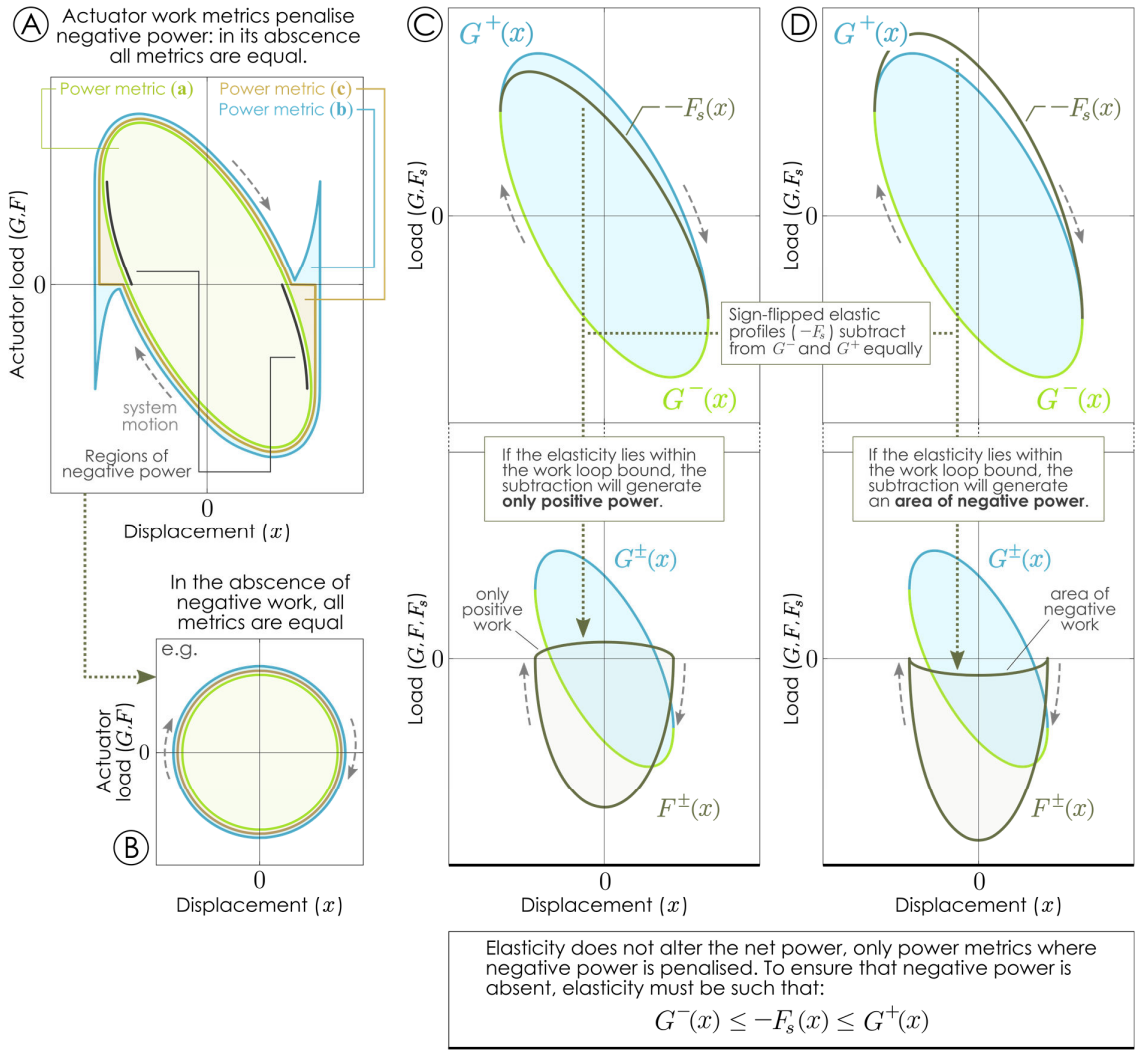

**Fig. 3: Explanation of the elastic-bound conditions in PEA systems**. (**B**) Work-loop interpretation of three of power metrics (a)-(c). The common feature of metrics (b) and (c) is that they penalise, in some way, regions of negative power; as such, they are particular forms of general metric (d). (**B**) In any situation, if the system work loop contains no negative power, then all metrics (a)-(d), are equivalent. (**C**) The sign-flipped elasticity, $-F_s(x)$, *i.e.*, the elastic reaction load profile, subtracts from the original work loop bounds, $G^{\pm}(x)$, equally, but leaves the system net work, metric (a), unchanged. As such, it may be seen that, if $-F_s(x)$ lies within original work loop bounds, then the system is free from negative power, and metrics (b)-(d) take minimal values. Or, (**D**), if at any point $-F_s(x)$ exceeds one of these bounds, then an area of negative power is created, and the actuator power consumption, under metrics (b)-(d), exceeds the net power. It follows that $G^-(x) \leq -F_s(x) \leq G^+(x)$ ensures minimum actuator power consumption.

**Step 2: Minimising actuator power consumption requires the absence of negative power.**
We then apply this logic to the case of actuator loads in inelastic, $G^{\pm}(x)$, and elastic, $F^{\pm}(x)$, versions of the same system, as per Eq. 7. We first observe, that, irrespective of elasticity, $F_s(x)$, both versions represent an identical net power – metric (**a**), the loop area. Physically, this is because elastic forces are energy-conserving; and geometrically, because $F_s(x)$ adds or subtracts from $F^+(x)$ and $F^-(x)$ equally (see Fig. 3B-3C). Thus, we can use elasticity to alter the actuator power consumption – metrics (**b**)-(**d**), in the elastic system – while keeping the net power, (**a**), constant. Given that metrics (**b**)-(**d**) can only ever be greater than or equal to the constant value of (**a**), the minimum possible value of metrics (**b**)-(**d**) is this constant value. This minimum will necessarily be achieved when there are no additional power requirements from actuator penalties, as per Fig 3A – that is, when the work loop contains no negative power. Thus, any elasticity that ensures that negative power is absent from the elastic system work loop, $F^{\pm}(x)$, will ensure that actuator power consumption is minimised, in all metrics (**b**)-(**d**).

**Step 3: Elastic bounding thus minimises actuator power consumption.**
Which elastic profile, $F_s(x)$, will ensure that the elastic work loop, $F^{\pm}(x)$, contains no negative power? In $F^{\pm}(x)$, negative power is generated when $F^+(x)$, becomes negative, or $F^-(x)$ becomes positive. We can see that elasticity, $F_s(x)$, functions as a simple additive between the inelastic and elastic work loops: $F^{\pm}(x) = G^{\pm}(x) + F_s(x)$, as per Eq. 7. To ensure that the upper curve, $F^+(x) = G^+(x) + F_s(x)$, is positive everywhere, it is required that $-F_s(x) \leq G^+(x)$. Similarly, to ensure that the lower curve, $F^-(x)$ is negative everywhere, $-F_s(x) \geq G^-(x)$ must hold. That is, to ensure that the actuator power consumption is minimised, the elastic-bound conditions of Eq. 10 must be satisfied. Figure 3B-3C illustrates this principle, with the effect of exceeding a bound indicated.

### 2.3. Optimality in series-elastic actuation (SEA) systems

As an alternative to PEA, series-elastic actuation (SEA) involves linking an elastic element to the system in series with the actuator (Fig. 4) – a configuration which generates behaviour analogous to that of classical base excitation [55]. The dynamics of a general nonlinear time-invariant 1DOF SEA system can be represented as:

$$D(x, \dot{x}, \ddot{x}, \dots) = F_s(u(t) - x(t)) = F(t), \qquad (11)$$

where $F(t)$ is the actuator load; $x(t)$ is the system displacement (or, kinematic output); $u(t)$ is the actuator displacement (or, kinematic input), and $F_s(\cdot)$ is the elastic profile, dependent on the relative displacement $u - x$. In SEA systems, load transmission, from actuator ($F$) to system dynamics ($D$), occurs instantaneously via elasticity ($F_s$), as per Eq. 11. This leads to several significant changes in behaviour vis-à-vis PEA systems. First, in SEA systems, the case of zero elasticity ($F_s = 0$) is not a physical actuation system: no load transmission is possible. Second, for a given kinematic output, $x(t)$, the load requirement, $F(t)$, is independent of elasticity and dependent only on the system dynamics, $D(\cdot)$ (Eq. 11). In contrast, the actuator displacement, $u(t)$, is dependent on elasticity, and thus links actuator power consumption to elasticity, via $P(t) = F(t)\dot{u}(t)$. It is nevertheless possible to derive an elastic-bound optimality condition for this system – a condition analogous to that of Eq. 10, but more complex. The process of obtaining this condition may be summarised in a four-step process:

**Step 1. Reformulate the system such that elasticity is additive**
Obtaining an elastic-bound condition – in which optimal elasticities are characterised via a set of bounds – relies on the additive properties of elasticity. In PEA systems, the system load ($F$) and elastic profile ($F_s$) are additive (Eq. 7); while in SEA systems, they are not. However, we may still construct an additive framework for elastic-bound analysis in SEA systems. Applying the inverse function of elasticity, $F_s^{-1}(\cdot)$, assuming it exists, to Eq. 11, we obtain:

$$u(t) = x(t) + F_s^{-1}(F(t)). \tag{12}$$

That is, the actuator displacement, $u(t)$ is the sum of the system displacement, $x(t)$, and the deformation of the elastic element, which is a function of the load across the element: $F_s^{-1}(F)$. This function, the inverse of elasticity, parallels the inverse relationship between stiffness and compliance or flexibility in linear systems [56]. This places a restriction on our analysis: only invertible, *i.e.*, monotonic, elasticities are permitted; non-monotonic (*e.g.*, bistable) elasticities are not. Note, also, that the case of zero elasticity in a PEA system is analogous to the case of infinitely rigid elasticity (zero compliance) in the SEA system: $x(t)$ represents an inelastic response, and $u(t)$ an elastic one (Fig. 4A).

**Step 2: Characterise system loops in displacement-load space**
As before, the system kinematic output, $x(t)$, is prescribed. It is periodic, with period $T$, and spans the displacement range $x(t) \in [x_1, x_2]$. Given that $F(t) = D(x, \dot{x}, \ddot{x}, \dots)$ as per Eq. 12, it follows that $F(t)$ is also prescribed, periodic, and spans some load range $F(t) \in [F_1, F_2]$. Instead of visualising these two periodic functions in load-displacement space ($F$-$x$, as per Section 2.2), the form of Eq. 12 leads us to visualise them in the rotated displacement-load space ($x$-$F$): in this space $F_s^{-1}(F)$ now has an additive effect. The loops traced out by $x(t)$ and $F(t)$ in $x$-$F$ are still work loops, for which area, $dF \cdot dx$, is work.

Following Section 2.2, we require, as a restriction on the analysis, that the work loop associated with the inelastic system dynamics, $x$-$F$, be a closed, simple curve, no more than bivalued at any $F \in [F_1, F_2]$, and representing net power dissipation (*i.e.*, the progression of time must represent *counterclockwise* travel around the closed loop). More specifically, we will require that the waveform $F(t)$ be composed of two half-cycles, not necessarily symmetric or in phase with $x(t)$, but each monotonic. If these conditions are fulfilled, then we can segment the inelastic work loop into upper, $X^+(F)$, and lower, $X^-(F)$, curves, associated with these separate, monotonic, half cycles of the force waveform, $F(t)$. The time-windows over which this segmentation occurs we denote $T^\pm$: they are the time windows associated with the half cycles of the force waveform, $F(t)$. As such, over each $T^\pm$, the force rate $\dot{F}(t)$, takes a particular consistent sign. $T^+ = \{t : \dot{F}(t) < 0\}$ is the window associated with the curve $X^+(F)$; and $T^- = \{t : \dot{F}(t) \geq 0\}$ is the window associated with $X^-(F)$. Using these time windows, we can segment and reparameterise a range of different variables in a consistent way. In an informal sense, formalised in the appendix (A.4), we can transform:

$$
\begin{aligned}
&\text{velocity:} & \dot{x}(t) \text{ into } \dot{X}^\pm(F), \\
&\text{force rate:} & \dot{F}(t) \text{ into } \dot{F}^\pm(F), \\
&\text{actuator disp.:} & u(t) \text{ into } U^\pm(F), \\
&\text{actuator velocity:} & \dot{u}(t) \text{ into } \dot{U}^\pm(F), \\
&\text{actuator power:} & P(t) \text{ into } P^\pm(F).
\end{aligned}
\tag{13}
$$

For instance, the values of $\dot{X}^+(F)$ are the values of $\dot{x}(t)$ at the times $t \in T^+$, associated with $X^+(F)$. For details on this process, see the appendix (A.4). Note $\dot{X}^\pm(F)$ and $\dot{F}^\pm(F)$ are already prescribed, by $\dot{x}(t)$ and $\dot{F}(t) = dF/dt$, respectively. Using the segmentations in Eq. 13, we can also formulate Eq. 12 in the $x$-$F$ domain, in a way that is directly analogous to the PEA system formulation (Eq. 7):

$$U^\pm(F) = X^\pm(F) + F_s^{-1}(F). \tag{14}$$

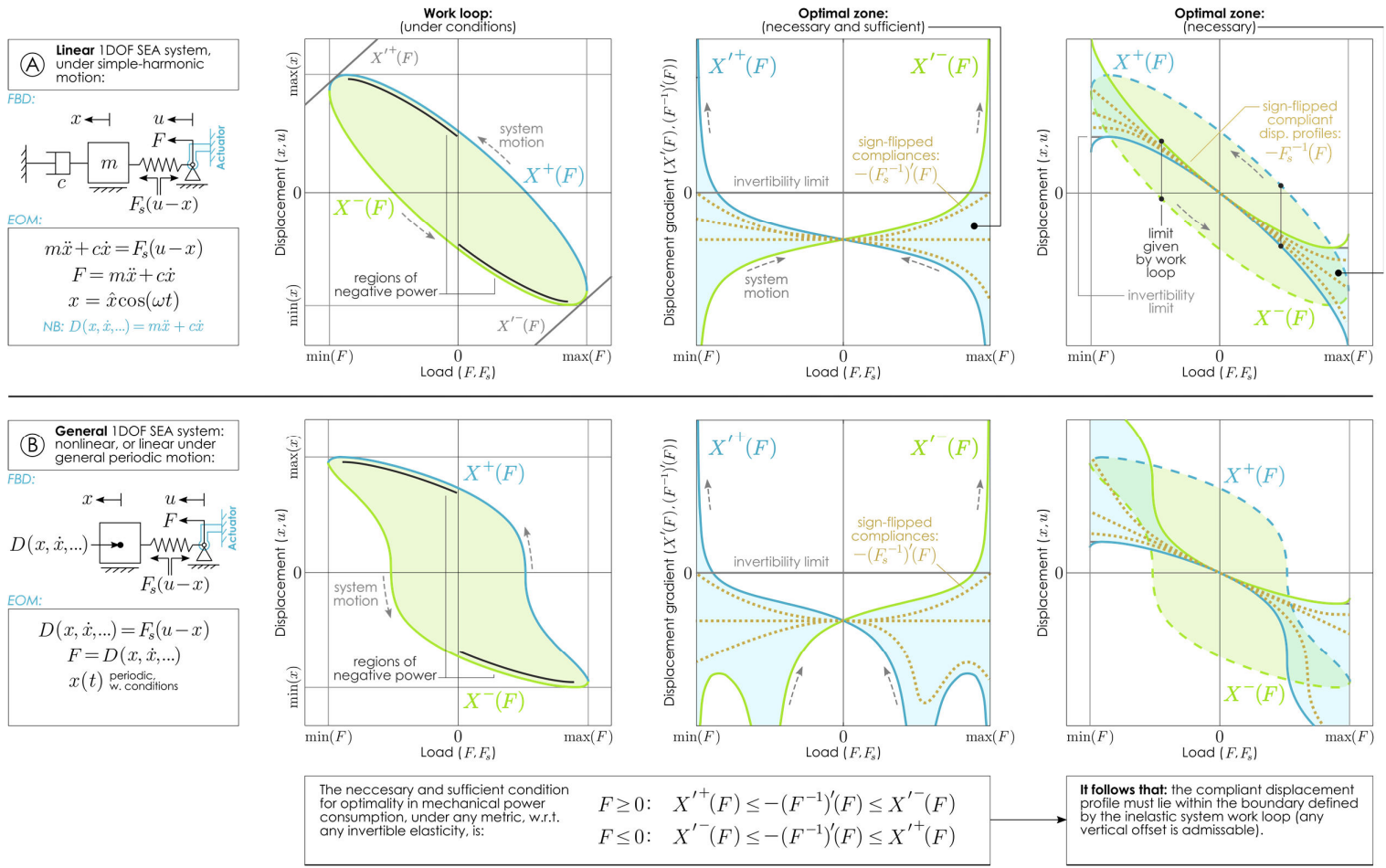

**Fig. 4.** Elastic-bound conditions in linear and nonlinear SEA systems. Free-body diagrams (FBD) and equations of motion (EOM) for (**A**) linear and (**B**) nonlinear SEA systems, with the elastic-bound conditions, Eq. 20, illustrated. Additional bounds, arising from the invertibility condition, are also illustrated, as well as the necessary (but not sufficient) direct elastic bound, Eq. 22.

**Step 3. Characterise actuator power consumption with respect to elasticity**

Despite the fact that area ($dx \cdot dF$ and $dF \cdot dx$) in both $F$-$x$ and $x$-$F$ space is directly equivalent, and represents mechanical work, the condition for negative power in the SEA system – loops $X^{\pm}(F)$ and $U^{\pm}(F)$ – is not as simple as in the PEA system. Returning to a time-based formulation of mechanical power, $P(t) = \dot{u}(t)F(t)$, we observe the derivative relation:

$$P(t) = \frac{d}{dt}\bigl(u(t)\bigr) \cdot F(t) = \frac{d}{dF}\bigl(u(t)\bigr) \cdot \dot{F}(t) \cdot F(t), \tag{15}$$

when $u(t)$ and $F(t)$ are differentiable. This relation transforms to the $x$-$F$ space as:

$$P^{\pm}(F) = U'^{\pm}(F) \cdot \dot{F}^{\pm}(F) \cdot F. \tag{16}$$

This expression for power consists of a multiplication of three terms: $F$, $\dot{F}$, and $U' = du/dF$. Differentiating Eq. 14 with respect to $F$, we obtain $U'^{\pm}(F)$ as:

$$U'^{\pm}(F) = X'^{\pm}(F) + (F_s^{-1})'(F), \tag{17}$$

where $(F_s^{-1})'(F)$ is the derivative of the inverse of the elastic load profile (i.e., the nonlinear compliance or flexibility profile of the elastic element[2]), and $X'^{\pm}(F)$ is the derivative of $X^{\pm}(F)$ with respect to $F$. $X'^{\pm}(F)$ may be computed directly from $X^{\pm}(F)$, or via:

$$X'^{\pm}(F) = \frac{\dot{X}^{\pm}(F)}{\dot{F}^{\pm}(F)}. \tag{18}$$

We have thus defined $P^{\pm}(F)$, Eq. 16, with the only remaining unknown being the system elasticity, expressed as $(F_s^{-1})'(F)$.

**Step 4. Derive elastic-bound conditions**

To obtain an elastic-bound condition for SEA systems, we must answer several questions. Firstly, does elasticity, $F_s(\cdot)$, alter the net power requirement, $\overline{P}_{(a)}$ (Eqs. 1, 8)? It does not: physically, elasticity is conservative, and mathematically, integrating Eq. 18 over the work loop causes $(F_s^{-1})'(F)$ to vanish. Secondly, what is the relationship between actuator power consumption metrics (**b**)-(**d**) and the net power, metric (**a**)? We may confirm again that metrics (**b**)-(**d**) are always greater than or equal to metric (**a**), and that the absence of negative work, $P(t) \geq 0, \forall t$, is the sole condition for these metrics to all be equal (the appendix, A.4). Finally, does the elasticity alter metrics (**b**)-(**d**)? In general, it does. It follows then, that any elasticity which ensures that negative power is absent from the system, necessarily minimises metrics (**b**)-(**d**) – i.e., is optimal in terms of actuator mechanical power consumption.

What, then, are the conditions on elasticity to ensure that negative power is absent from the system? By Eqs. 15-16 negative power ($P^{\pm}(F) < 0$) is generated when the signs of the force ($F$), force rate ($\dot{F}$), and gradient of the resultant work loop ($U'$), all multiply to be negative. A graphical illustration of this is given in Fig. 4. Note that this condition may be confirmed by rotating the loop back to load-displacement space ($F$-$u$) and observing where $P = F\dot{u}$ changes sign. To ensure only positive power is present, $P^{\pm}(F) \geq 0, \forall F \in [F_1, F_2]$, we substitute Eq. 17 into Eq. 16 and obtain:

$$-(F_s^{-1})'(F) \cdot \dot{F}^{\pm}(F) \cdot F \leq X'^{\pm}(F) \cdot \dot{F}^{\pm}(F) \cdot F, \qquad \forall F \in [F_1, F_2]. \tag{19}$$

---

[2] An example: for a linear elasticity, we have the elastic load profile $F_s(\delta) = k\delta$; the compliant displacement profile $F_s^{-1}(F) = F/k$; and the (constant) compliance profile $(F_s^{-1})'(F) = 1/k$.

Accounting for all possible combinations of signs for $F$ and $\dot{F}$, as we do in detail in the appendix (A.4), we obtain a set of conditions on $(F_s^{-1})'(F)$. These are the SEA elastic-bound conditions:

$$\begin{aligned} \text{for } F_1 \leq F \leq 0 \quad &: \quad X'^{-}(F) \leq -(F_s^{-1})'(F) \leq X'^{+}(F); \\ \text{for } 0 \leq F \leq F_2 \quad &: \quad X'^{+}(F) \leq -(F_s^{-1})'(F) \leq X'^{-}(F), \end{aligned} \quad (20)$$

which are sufficient for energetic optimality, in all metrics (**b**)-(**d**), under the system conditions that have been outlined (invertible elasticity, relevant curve conditions). This concludes the four-step process for formulating the elastic-bound optimality condition in this SEA system – a full proof, as well as a derivation of $X'^{\pm}(x)$ in a specific case, is presented in the appendix (A.2, A.4). In intuitive terms: negative power is generated when $U'$ reverses direction within a half-cycle, as this represents a reversal in $\dot{u}$ ($P = F\dot{u}$). To eliminate this negative power, we require that the gradient of the compliance, $(F_s^{-1})'$, is bounded by $X'$, because these three gradient terms show an additive relationship, $U' = X' + (F_s^{-1})'$. Figure 4 shows a visualisation of these elastic-bound conditions, in general linear and nonlinear SEA systems, alongside several relevant notes, as follows:

**Note (i)**. Eq. 20 indicates that any invertible nonlinear compliance profile, $(F_s^{-1})'(F)$, which, under a sign-flip, lies within the bounds of the inelastic work loop gradient, $X'^{\pm}(F)$, is optimal in mechanical power consumption – irrespective of the metric of mechanical power. As illustrated in Fig. 4, the requirement for invertible elasticity necessitates that the compliance be either always-stable, $(F_s^{-1})'(F) \geq 0, \forall F \in [F_1, F_2]$, or always-unstable, $(F_s^{-1})'(F) \leq 0, \forall F \in [F_1, F_2]$, with no more than a single equilibrium. The choice, if any, between the two forms of stability is given by the nature of the bounds, $X'^{\pm}(F)$, in Eq. 20 – but for the systems that we are studying, the always-stable case is the only relevant case.

**Note (ii)**. Eq. 20, as stated, is sufficient but *not* necessary for optimality in mechanical power. Eq. 20 becomes necessary if the inelastic system itself, $x(t)$, $D(x, \dot{x}, \ddot{x}, ...)$ and $\dot{D}(x, \dot{x}, \ddot{x}, ...)$, satisfies a set of further conditions:

$$\begin{aligned} X'^{+}(F) &\leq X'^{-}(F), \quad \forall F \in [F_1, 0], \\ X'^{-}(F) &\leq X'^{+}(F), \quad \forall F \in [0, F_2], \\ \text{and, } X'^{+}(0) &= X'^{-}(0). \end{aligned} \quad (21)$$

Eq. 20 represents the conditions that must be satisfied in order for the actuator power requirement, $P(t)$, to contain no regions negative power – if this is satisfied, the system is energetically optimal with respect to elasticity. However, some systems, it may not be possible to satisfy Eq. 20 with any elasticity – a state of no negative work may not be accessible, and power metrics (**b**)-(**d**) may *always* be greater than metric (**a**). Eq. 21 gives the conditions on the inelastic system such that this state is guaranteed to be accessible – and thus, Eq. 20 is the necessary and sufficient condition for optimality in actuator mechanical power consumption. In systems not satisfying Eq. 21, it is still possible to optimise actuator power consumption with respect to elasticity, but this optimal state will still contain regions of negative power, and Eq. 20 will not be satisfied. This accessibility and non-accessibility of optimal states without negative power is a fascinating area that warrants further investigation.

**Note (iii)**. Eq. 20 is an inequality condition: in all admissible (and non-trivial) systems, it defines a continuous range of optimal compliance profiles. These optimal compliance profiles show a common feature: Eq. 20 functions as an equality condition at zero load ($F = 0$), implying that the compliance at $F = 0$, $(F_s^{-1})'(0)$, must take the value common to $-X'^{\pm}(0)$ (Eq. 22), to ensure energetic optimality via the absence of negative work, as per note (**ii**). If the system satisfies the conditions of note (**ii**), then the optimal elasticities for this given system are united by this common condition: the stiffness of the elastic element about its single rest state, *i.e.*, the stability of its single equilibrium.

**Note (iv).** Eq. 20 is a condition is on the compliance of the elastic element, $(F_s^{-1})'(F)$: the gradient of the compliant displacement profile, not the compliant displacement profile itself. Any offset is permissible, *i.e.*, any $C$ in $F_s^{-1}(F) = \int (F_s^{-1})'(F) \, dF + C$, though such an offset is not useful: it merely represents a static offset in the actuator displacement, $u(t)$, with the actuator displacement about the system equilibrium, *i.e.*, $u(t) - F_s^{-1}(0)$, remaining unaltered.

**Note (v).** Integrating Eq. 20 yields a condition on the compliant displacement profile itself, $F_s^{-1}(F)$, which gives its maximum possible range – in the format of a bound like Eq. 7. The condition is:

$$X^+(F) - X^+(0) \leq -F_s^{-1}(F) - F_s^{-1}(0) \leq X^-(F) - X^-(0), \tag{22}$$

where $X^{\pm}(0)$ and $F_s^{-1}(0)$ account for the fact that $F_s^{-1}(0)$ can equal any constant, as per (**iv**). Fig. 4 illustrates this condition: to visualise the range of compliant displacement profiles, $F_s^{-1}(F)$, that the condition on $(F_s^{-1})'(F)$ represents, we can integrate the derivative bounds, $X'^{\pm}(F)$, back into conditions on $X^{\pm}(F)$. If we select the same integration constants for both bounds, $X^-(0) = X^+(0)$, then we can interpret the resulting bounded area as the maximum possible range of the compliant displacement profile (for a given constant equilibrium location). Eq. 22 is necessary, but not sufficient, for energetic optimality: the gradient condition, Eq. 20, must still be satisfied within these bounds.

### 2.4. Invariant metrics

Elasticities that satisfy the elastic-bound conditions of Sections 2.2-2.3 show several additional properties. In particular, across the set of optimal elasticities, there is a wide range of invariant metrics: not only the net power, metric (**a**) (independent of elasticity) and power consumption metrics (**b**)-(**d**) (invariant by nature of their optimality), but others. These additional invariant metrics are highly relevant, because several represent additional, system-specific metrics of power consumption – *e.g.*, metabolic, or electrical power consumption, describing processes deeper within the system drivetrain (Fig. 1).

For instance: in the PEA system, the actuator displacement, $x(t)$, is independent of elasticity. In the SEA system, the actuator load, $F(t)$, is independent of elasticity. Thus, in the PEA system, any metric based solely on $x(t)$ is invariant, and in the SEA system, any metric based solely on $F(t)$ is invariant. The latter includes the following two sets of power-consumption metrics. (**i**) Resistive electrical power losses, measured via the force-squared metric [43]:

$$\overline{P}_{F^2} \propto \int_0^T F(t)^2 \, dt. \tag{23}$$

In a range of electromechanical actuators, including direct-current (DC) motors and voice-coil actuators (VCA) [57, 58], the proportionality of force to current ($F \propto I$) and resistive power loss to the square of current ($P \propto I^2$) leads to Eq. 23. (**ii**) Isometric muscular power consumption, measured via a range of force-based metrics, including $\overline{P}_{F^2}$ and also [42, 45–47]:

$$\overline{P}_{|F|} \propto \int_0^T |F(t)| \, dt, \quad \overline{P}_{|\dot{F}|} \propto \int_0^T |\dot{F}(t)| \, dt, \quad \overline{P}_{\dot{F}^2} \propto \int_0^T \dot{F}(t)^2 \, dt. \quad (24)$$

In the SEA system, all these power metrics are invariant with respect to elasticity – that is to say, the choice of elasticity can be made without reference to how these metrics will be affected. Of course, other factors (*e.g.*, the desired system output kinematics) will still affect these metrics, and must be considered in a complete optimisation of system power requirements.

Compared to SEA systems, PEA systems less well suited to optimising complete-system power consumption (*i.e.*, when load integral metrics, such as $\overline{P}_{F^2}$, are included). In general, PEA systems do not ensure the invariance of load-based metrics with respect to elasticity – however, there is one key exception. In PEA systems, the absolute force metric $\overline{P}_{|F|}$ is invariant across the space of elastic functions satisfying the elastic-bound conditions, *provided* that the prescribed kinematic waveform, $x(t)$, is composed of two symmetric half-cycles. That is, $x(t) = x(T - t), \forall t$. This condition is satisfied for a range of kinematic profiles typically used in forced-oscillation applications: *e.g.*, sine waves, smoothed triangle waves, and smoothed square waves. It is only violated under less typical kinematic profiles, *e.g.*, a smoothed sawtooth wave. A proof of this constant optimality is given in the appendix (A.5). This exception to the general non-invariance of load integral metrics mean that PEA systems may still be well-suited to minimising overall power consumption in muscular systems where the metric $\overline{P}_{|F|}$ is relevant [59].

## 2.5. Is this behaviour resonant?

Having established the conditions for energetic optimality, with respect to elasticity, in the PEA and SEA systems that we have studied, a natural question arises: to what extent do these energetically optimal states represent nonlinear resonant states? Nonlinear resonance, as a broad phenomenon, is characterised in different ways in different contexts: in terms of global resonance [27–30], coherence resonance [60–62], the jump phenomenon [63–65], or resonant wave interaction [66]. The states that we describe are related to a global resonance (*i.e.*, energetic) characterisation of nonlinear resonance. These states are the (concurrent) maxima of a family of energy-based nonlinear transfer ratios: the metric ratios, $\overline{P}_{(a)}/\overline{P}_{(i)}$, $i \in \{b, c, d\}$, and also the input-output power transfer ratios ($H$):

$$\text{PEA: } H = \frac{\int_0^T |G\dot{x}| \, dt}{\int_0^T |F\dot{x}| \, dt}, \quad \text{SEA: } H = \frac{\int_0^T |F\dot{x}| \, dt}{\int_0^T |F\dot{u}| \, dt}. \quad (25)$$

These transfer ratios represent the ratio of system output power to system input power, in absolute terms. For instance, for the PEA system, the denominator term $|F\dot{x}|$ represents the power provided by the actuator (the input); and the numerator term $|G\dot{x}|$, $G = D(x, \dot{x}, \ddot{x}, \dots)$, represents the power dissipated to the environment (the output). Variant forms, *e.g.*, using a positive-only power measure, are also available.

In general, these nonlinear transfer ratios are extensions of power-, energy- and work-based transfer functions in linear systems, as used *e.g.*, in vibroacoustics [67, 68], seismic base excitation [69–71] and energy harvesting [72, 73]. By the conservation of energy, the maximum possible value that these ratios can take, under any condition, is unity. The condition for obtaining this maximum of unity is the global resonance condition of Ma and Zhang [27–30], *i.e.*, the absence of negative power. This condition describes states in a nonlinear system which constitute a generalisation of the energetic properties of linear resonance: perfect energy transfer through the system. In physical terms, these nonlinear resonant states show complete inertial power absorption: the power consumption of inertial effects are completely absorbed by elasticity, with only the power consumption of dissipative effects remaining. The relationship between these states and those characterised by other forms of nonlinear resonance – *e.g.*, coherence resonance [60–62] – is currently unclear, and in differing systems, the characterisation most appropriate to the system design objective is likely to differ. However, for deterministic systems in which energetic efficiency is paramount, our generalisation of the energetic properties of linear resonance is appropriate. We will now study some of these systems in more detail.

## 3. Systems-design applications

### 3.1. Motivation for exploring the elastic-bound conditions

The elastic-bound conditions of Sections 2.2-2.3 provide a methodology for designing an elastic element within an actuated system to minimise the mechanical power consumption of the actuator under resonant conditions. If any system, undergoing any periodic motion, is admissible under the analysis conditions, then the available range of optimal elasticities can be constructed. This range of optimal elasticities may encompass many different forms of elasticity behaviours: *e.g.*, linear, bistable, strain-hardening, strain-softening, freeplay, and others. Even within the bounds of energetic optimality, the choice of elastic behaviour is a further design consideration: it allows different forms of behaviour to be designed into the system (Section 3.2). In making this choice, there are several factors to be weighed, including:

**Factor (i)**. The choice of elasticity allows a selection of different actuator waveforms: load (PEA) or displacement (SEA) waveforms, as required to generate the prescribed, nominal output kinematics. Each of these waveforms will represent an identical mechanical power consumption, but the difference in waveform may make them suitable for different actuator types – for instance, actuator suited to a low duty cycle operation.

**Factor (ii)**. The choice of elasticity allows a selection of different system behaviours under off-nominal conditions – *i.e.*, different output behaviour for any off-nominal input waveform; and a different transient response. This allows a wide range of behaviour to be introduced into the system, while maintaining the energetic optimality in the nominal state. As a subclass of this capability, the choice of elasticity allows control over the system response to quasistatic loading; and the nature and location of the system equilibria (for instance, a linear elasticity with one stable equilibrium, *vs.* a bistable elasticity with one unstable and two stable equilibria).

**Factor (iii)** Not all relevant metrics of system performance, efficiency and capability requirement remain optimal, or invariant, under the choice of elasticity – even at the nominal state. For instance, in a PEA system, the nominal electrical resistive losses ($\overline{P}_{F^2}$), nominal peak load requirements ($\max|F|$), and nominal peak power requirements ($\max|P|$), are not invariant. The latter two affect the system's peak capability requirements, and the former contributes to the system's overall efficiency. These metrics may thus limit use of elasticity to design particular forms of behaviour into the system, as per factor (**ii**). In contrast, the SEA system is significantly simpler: the peak power requirement ($\max|P|$) is the only immediately relevant metric which is non-invariant. But, because the actuator load waveform is invariant, it cannot be altered through the choice of elasticity, as per factor (**i**). This distinction between the properties of PEA and SEA systems may be a relevant design consideration.

As can be seen, while there are multiple competing factors that must be weighed in the choice of elasticity, there is also significant capability for altering whatever system properties are prioritised by the designer – all while keeping the mechanical efficiency of the system (at the very least) at an optimal state. Over the following two sections, we will detail some of the choices that can be made, and their implications for an extensive range of different systems.

**3.2. PEA systems-design applications**
As noted in Section 3.1, in PEA systems, the choice of elasticity alters the actuator load requirement associated with the nominal output kinematics. This has the disadvantage, as per Section 2.4, of not maintaining invariance in metrics dependent solely on actuator load (*e.g.*, electrical resistive losses); but the advantage of allowing particular actuator load requirement waveform properties to be designed into the system via elasticity. In addition, our PEA analysis framework also permits non-invertible elasticities (*e.g.*, bistable elasticities, see Section 2.2), allowing significant control over the system equilibria and quasistatic behaviour. Figure 5 describes several forms of optimal elasticity in a general PEA system, and their implications for systems design. These forms are shown for a PEA system with linear inertia and linear dissipation, undergoing simple harmonic motion around a zero-valued reference ($x = 0$); the construction of $G^\pm(x)$ for such a system is given within the appendix (A.1)

**Form 1. Freeplay elasticities**
Freeplay elasticities are characterised by a region of zero elastic force – the freeplay region – often located around a rest state. For instance, the elastic force profile:

$$F_s(x) = \begin{cases} -kx & x \leq -\delta, \\ 0 & -\delta \leq x \leq \delta, \\ kx & x \geq \delta, \end{cases} \quad (26)$$

represents a freeplay elasticity with linear stiffness $k$ and freeplay region $-\delta \leq x \leq \delta$, for some $\delta \geq 0$. Freeplay elasticities commonly arise from backlash (*i.e.*, clearance and tolerance) effects in rigid mechanisms such as gear trains [74], and are typically unintentional and undesired. Nevertheless, such elasticities can still be optimal under the elastic-bound condition of Section 2.2. Fig. 5A shows an example optimal freeplay elasticity. In general, the elastic-bound condition implies that a wide range of different freeplay elasticities are capable of generating an energetically-optimal (*i.e.*, global-resonant) state. For instance, if we define the critical displacements $l^\pm$ such that $G^+(l^+) = 0$ and $G^-(l^-) = 0$, then any freeplay, as per Eq. 26, with $0 \leq \delta \leq \min\{-l^+, l^-\}$, can be optimal, with the correct $k$. Or, in practical terms, an energetically optimal resonant response can be achieved even if an unintentional freeplay ($\delta$) is present in the system, provided $0 \leq \delta \leq \min\{-l^+, l^-\}$. All that is required is an alteration of the actuator load waveform, as illustrated in Fig. 5A.

Going further, however, there are also cogent reasons to consider the intentional use of freeplay elasticities within PEA systems. For instance, consider a system designed for two modes of operation – a high-frequency mode of operation, and a quasistatic mode of operation. Such a system could represent several forms of biomimetic propulsion and/or flow control, *e.g.*: (**i**) a legged robot [24, 40] designed both for running and for slow, careful, walking; (**ii**) an aquatic robot [75, 76] designed for both rapid and slow swimming; or (**iii**) an active gurney flap in an aerofoil or hydrofoil [77, 78], designed for both oscillatory flow control and slow flight control. Applying the elastic-bound conditions to the high-frequency mode of operation, we obtain a range of elasticities minimising the mechanical power consumption of this high-frequency state. However, some of these elasticities (*e.g.*, the resonant linear elasticity) introduce a problem: during the quasistatic mode of operation the actuator must actively work against this resonant elasticity. This quasistatic resistance effect might increase the maximum load capability of the actuator (*i.e.*, necessitating a stronger actuator than otherwise) and it might degrade the efficiency of the quasistatic mode of operation (*e.g.*, via the energetic costs of quasistatic actuator loading). A freeplay elasticity – designed, for instance, according to the schematic of the contact-based freeplay mechanism (Fig. 5A) – provides one solution to this problem: a region over which the elasticity exerts no load, leaves the quasistatic mode of operation in an entirely inelastic state: theoretically, with zero load requirement, but in practice, with small inertial and dissipative loads. The freeplay elasticity has altered this quasistatic mode of operation from one would have required significant (linear elastic) load, to one which theoretically requires zero load: a significant alteration. The region (in $x$) of zero load cannot span the full kinematic range of the high-frequency mode of operation, but in the applications described, this may be sufficient to ensure, *e.g.*, efficient slow walking or swimming. A freeplay elasticity thus allows the efficiency and performance of the quasistatic mode of operation to be increased, at no cost to the mechanical power consumption of the high-frequency mode of operation[3].

**Form 2. Bistable elasticities**
Bistable elasticities consist of an unstable equilibrium bracketed by stable equilibria – Fig. 5A illustrates an energetically optimal bistable elasticity in a PEA system with linear inertia and dissipation. Naturally, availability of energetically optimal bistable (and multistable) elasticities allows a range of different system equilibria to be designed into the system at no energetic cost. For instance, bistable elasticities with stable equilibria at high displacement (Fig 5A) may be useful for systems designed to oscillate outwards from a retracted equilibrium, including aerofoil vibrational active flow control devices such as active gurney flaps [77, 78]: in such devices, an equilibrium located near an extremum of the oscillatory displacement range may represent a streamlined aerofoil profile. Structurally, bistable elasticity can be implemented via the bistable mechanism, based on two monotonic elasticities, illustrated as a schematic in Fig. 5A.

But again: there are further reasons to consider the use of bistable elasticities for optimising the energetic behaviour of PEA systems. Several actuator types, including solenoids and shape memory alloys, are suited to low duty-cycle operation – actuation during only a small window of the load cycle [79–82]. In such actuators, there may be an imperative to concentrate loads in particular regions of the oscillatory cycle, and leave other regions load-free. A bistable elasticity tuned to lie along the boundary of the inelastic system work loop (*i.e.*, tuned to both the inertia and the dissipation of the system) can yield actuator load requirement waveform with a duty cycle down to 50% (Fig. 5A) or even below[4]. For instance, in the system illustrated in Fig. 5A (with linear inertia and dissipation, and undergoing simple harmonic motion) we can construct the family of bistable elasticities:

---

[3] In an electromechanical system, resistive losses ($\overline{P}_{F^2}$) may increase; but in biological systems, there may be no such penalty, because $\overline{P}_{|F|}$ may be invariant (Section 2.4). This may provide an explanation for strain-hardening nonlinearities in, *e.g.,* the insect flight motor [12].

[4] These elasticities lead to a limit state in which the actuator is at zero load over 50% of the *work loop* (*cf.* Fig 5A). If different quarter-cycles of the work loop are (kinematically) faster than others, then the limit state of the time-domain duty cycle can be <50%.

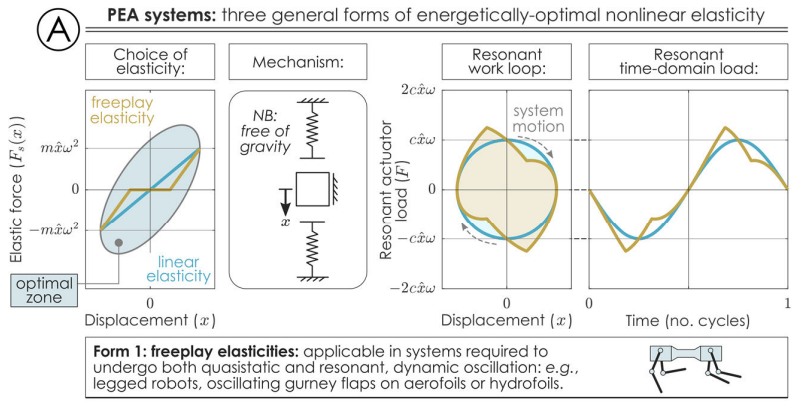
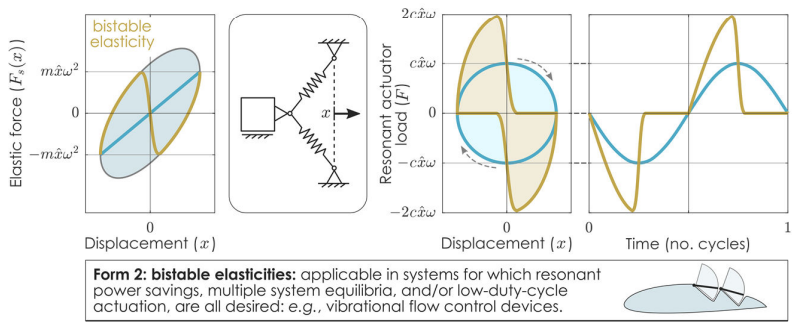
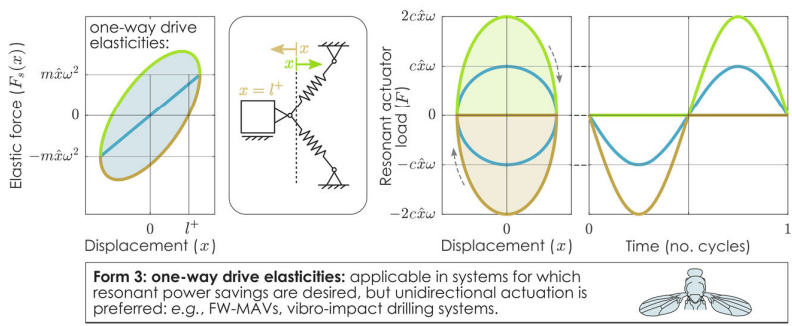
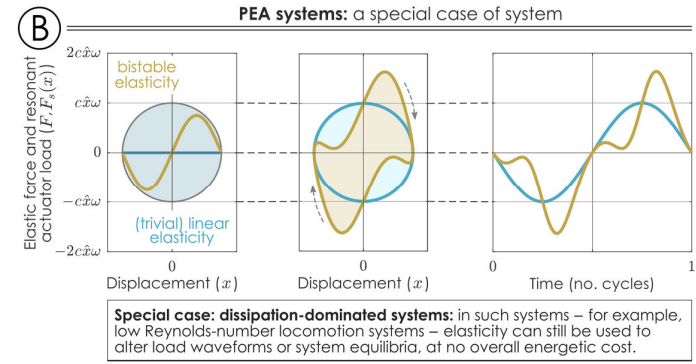
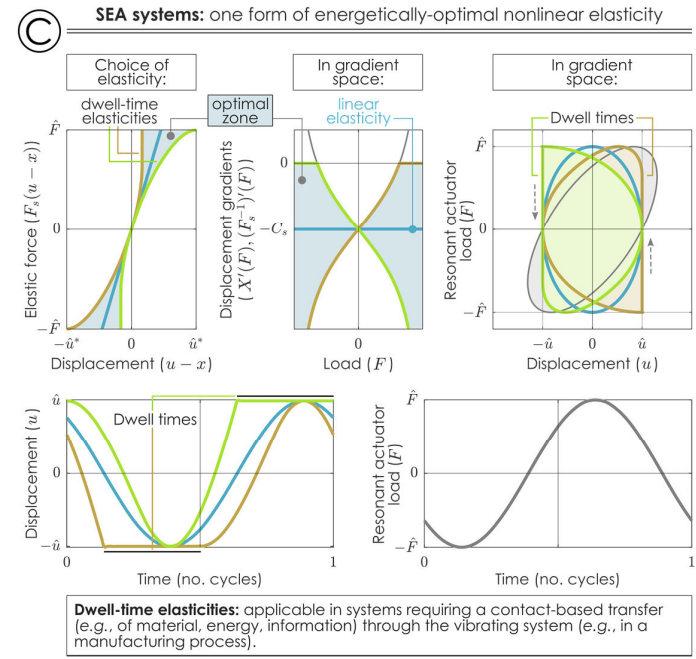

Page 19 of 42

**Fig. 5: Forms of optimal nonlinear elasticity in PEA and SEA systems, and their implications for systems design.** (**A**) Three forms of optimal nonlinear elasticity in a linear PEA system; (**B**) The special case of a dissipation-dominated PEA systems; (**C**) One form of optimal nonlinear elasticity in a linear SEA system. In all cases, the nonlinear elastic profile is illustrated (without a sign flip), and compared to an optimal linear elasticity. Resonant work loops and time-domain actuator loads, under the linear and nonlinear elasticity, are compared. Example applications are briefly discussed. Where particularly relevant, a simple mechanism, capable of generating the required elasticity, and composed only of elements with monotonic elasticity, is illustrated. Note also the range of scaling parameters used to characterise the SEA system: $\hat{u} = \hat{x}c/\sqrt{m^2\omega^2 + c^2}$; $\hat{u}^* = \hat{x}(m\omega + c)/\sqrt{m^2\omega^2 + c^2}$; $\hat{F} = \omega\hat{x}\sqrt{m^2\omega^2 + c^2}$ and $C_s = m/(m^2\omega^2 + c^2)$ (*cf.* the appendix, A.2)

$$F_s(x) = \begin{cases} -G^-(x) & x \leq -\delta, \\ G^+(\delta)\dfrac{x}{\delta} & -\delta \leq x \leq \delta, \\ -G^+(x) & x \geq \delta, \end{cases} \quad (27)$$

controlled by the parameter $\delta$. For $0 < \delta \leq \max(x)$, this elasticity satisfied the elastic-bound conditions of Section 2.2. For $0 < \delta < l^+$, where $l^+$ are such that $G^+(l^+) = 0$, this elasticity is bistable; for $\delta = l^+$, this elasticity is a freeplay strain-hardening elasticity; for $l^+ < \delta < \max(x)$, the elasticity is strain hardening; and for $\delta = \max(x)$ the elasticity is linear. In the bistable regime, $0 < \delta < l^+$, the duty cycle of the actuator load requirement waveform reaches 50% in the limit $\delta \to 0$. While this limit is not practically achievable, operating near this limit – as in the elasticity shown in Fig. 5A, which represents a smoothed version of Eq. 27 – leads to similarly low duty cycles. This allows the construction of energetically optimal resonant drive systems using actuators which are only capable of providing low duty-cycle load waveforms, *i.e.*, concentrated load in only one area of the load cycle. There is, however, an even more powerful form of duty-cycle minimisation in this system, as follows.

**Form 3. The one-way drive elasticities**
In any PEA system that is admissible under our analysis conditions (Section 2.2) a pair of unique elasticities ($F_{s,1}$ and $F_{s,2}$) necessarily exist. They can be defined simply:

$$\begin{aligned} F_{s,1}(x) &= -G^+(x), \\ F_{s,2}(x) &= -G^-(x); \end{aligned} \quad (28)$$

that is, they represent the upper and lower boundaries of the sign-flipped work loop, $-G^\pm(x)$, and necessary satisfy the elastic-bound condition, Eq. 10. In physical terms, the nature of these elasticities is system-dependent: in the system illustrated in Fig. 5A, one implementation could involve bistable elasticities operating to one side of an unstable equilibrium (*i.e.*, a stable equilibrium that starts to destabilise in one direction). An equivalent mechanism, based on two monotonic elasticities, is illustrated in the schematic alongside[5].

---

[5] These elasticities are not always exactly realisable – *e.g.*, in Fig. 5A, the system's linear dissipation necessitates an infinite unstable stiffness at the displacement extrema. However, there are other systems in which they are exactly realisable: *e.g.*, in systems with quadratic dissipation, where $\max_x|\partial_x G^+(x)|$ is finite, *cf.* the appendix (A.1).

The elasticities defined by Eq. 28 are united by several unique properties. By their definition, they leave one boundary of the resonant work loop at a state of zero load. As per Eq. 7:

$$F^{\pm}(x) = G^{\pm}(x) + F_s(x),$$
$$F_{s,1}(x) = -G^+(x) \therefore F^+(x) = 0,$$
$$F_{s,2}(x) = -G^-(x) \therefore F^-(x) = 0. \tag{29}$$

The other boundary of the work loop lies at a state of unidirectional load:

$$F_{s,1}(x) = -G^+(x) \therefore F^-(x) = -2G_{\mathrm{arc}}(x),$$
$$F_{s,2}(x) = -G^-(x) \therefore F^+(x) = 2G_{\mathrm{arc}}(x). \tag{30}$$

where $G_{\mathrm{arc}}(x) \geq 0, \forall x$. That is to say, the actuator load requirement for this system is now unidirectional, and the system remains energetically optimal. Forces in only one direction are required for resonant excitation – in the case of the system illustrated in Fig. 5C, in the waveform of a half-sine wave – and for this reason, we term these elasticities the one-way drive elasticities. And if the property of unidirectional actuation was not enough, the one-way drive elasticities additionally guarantee that the actuator load requirement duty cycle is exactly 50% or lower – by nature of the fact that one boundary of the work loop is left a state of zero load[6]. The implications of the one-way drive elasticities are extensive. They allow a single unidirectional actuator (*e.g.*, a solenoid or combustion cylinder) to generate energetically-optimal resonant oscillations – as an additional benefit, the actuator duty cycle is, at maximum, 50%. The principle is intuitive, if surprising: an elastic element can be devised which optimally recreates one half-cycle of the oscillatory motion, via stored energy from the other half cycle.

Applications for the one-way drive elasticities are widespread: engineering systems undergoing forced resonant oscillation are, almost universally, subjected to bidirectional actuation, as a necessary condition of linear resonance. In systems in which the damping is sufficiently well known that the work-loop boundary elasticities can be accurately defined, the possibility of unidirectional resonant actuation significantly simplifies actuation requirements, and allows new classes of actuator to be used, without duplication: solenoids, unidirectional hydraulic actuators, combustion cylinders, and even reaction engines. More specific applications include FW-MAVs and vibro-impact drilling systems, both of which stand to benefit from a reduction in actuator complexity.

**Special case: Dissipation-dominated systems**
Finally, a fascinating special class of behaviour arises not with a particular form of elasticity, but in particular PEA systems. The systems in question are dissipation-dominated: with zero effective inertia and no elasticity in the original dynamics, *i.e.*, where $D(x, \dot{x}, \ddot{x}, \ldots) \approx D(\dot{x})$ with $D(0) = 0$. Figure 5B illustrates one such system: a system identical to that portrayed elsewhere in Fig. 5, but with zero inertia. In inertialess systems, no reduction in actuator power consumption is available via the introduction of elasticity. But energetically-optimal (that is, energetically-neutral) elasticities can still be introduced into the system in order to alter the nominal actuator load requirement waveform, and the system behaviour under off-nominal conditions (*e.g.*, quasistatic behaviour, equilibria stability and location).

---

[6] If the kinematic waveform, $x(t)$, is composed of two symmetric half-cycles, *i.e.*, $x(t) = x(T - t), \forall t$, then both one-way drive elasticities will generate load requirement waveforms of duty cycle 50% (Fig. 5A). If this is not the case, then one will generate a waveform of duty cycle <50%, and the other, a waveform of duty cycle >50%.

The principle for constructing these energetically-neutral elasticities is identical to that already described. Bistable elasticities can be constructed in order to alter the system equilibria and their stability, and also reduce the actuator duty cycle. One-way drive elasticities can be constructed in order to reduce the actuator load requirement to a unidirectionality, and also reduce the actuator duty cycle. Simple (*e.g.*, piecewise linear) freeplay elasticities reduce to the trivial case of zero elasticity; but more complex freeplay and/or contact-based elasticities can be constructed. For instance, we can conceive of a destabilizing elasticity with a single stable equilibrium and two unstable equilibria (Fig. 5B). Such an elasticity provides a way to construct a high-stability equilibrium in a dissipation-dominated system, while maintaining energetic neutrality. Such a system may have applications in low Reynolds-number locomotion [8, 9, 83]. Furthermore, the principle of energetically-neutral load waveform alteration provides competing explanations for elastic elements found in biological propulsion systems (*e.g.*, the insect flight motor, and low-Reynolds number locomotion). The evolutionary motivation for these structures could be the synchronisation of muscular load capability with propulsion load requirements, alongside or instead of any overall energy savings. This alternate role of elasticity in such systems has not previously been considered.

### 3.3. SEA systems-design applications

Comparing SEA to PEA systems – in terms of the choice of optimal elasticity, and its effects – we note the following advantages and disadvantages. Advantage (**i**): SEA systems have the significant advantage of retaining invariance, with respect to elasticity, in the actuator load requirement waveform – and thus also any purely load-based metrics, *e.g.*, electrical resistive loss, and peak load requirement. Disadvantage (**i**): as a necessary corollary, elasticity cannot be used to alter the load requirement waveform. Low duty-cycle or one-way drive systems cannot be constructed. Disadvantage (**ii**): our analysis does not permit non-invertible (*e.g.*, bistable) elastic profiles in SEA systems. Only elasticities with a single stable equilibrium (within the oscillatory displacement range) are permitted. Disadvantage (**iii**): the elasticity at the system's (single) equilibrium point, is prescribed. Under Eq. 20: $-(F_s^{-1})'(0) = X'^{+}(0) = X'^{-}(0)$ for energetic optimality. This condition is illustrated in Figs. 4 and 5: as can be seen, at $F = 0$ (the equilibrium), a particular fixed gradient is required. This limits the types of elastic behaviour that can be designed into an SEA system. Overall, none of the types of elastic behaviour studies in Section 2.3 are available in the SEA system – but there is one differing form of elasticity which is worth noting.

**The dwell-time elasticities**

As in the PEA system, elasticities in the SEA system that lie along the boundary of the optimal zone (Eq. 20) show special properties. In the SEA system, these special properties are manifested not in load, $F(t)$, which is invariant, but in actuator displacement, $u(t)$. For instance, consider the elasticities given by:

$$(F_s^{-1})'(F) = -X'^{+}(F), \quad \text{so, } e.g. \quad F_s^{-1}(F) = -X^{+}(F), \tag{31}$$
$$(F_s^{-1})'(F) = -X'^{-}(F), \quad \text{so, } e.g. \quad F_s^{-1}(F) = -X^{-}(F),$$

as shown in Fig. 5C. These elasticities ensure that one side of the actuator work loop, $U^{\pm}(F)$, lies at a state of zero displacement – that is:

$$U^{\pm}(F) = X^{\pm}(F) + F_s^{-1}(F),$$
$$F_s^{-1}(F) = -X^{+}(F) \therefore U^{+}(F) = 0, \tag{32}$$
$$F_s^{-1}(F) = -X^{-}(F) \therefore U^{-}(F) = 0.$$

In other words, for approximately one half-cycle of its oscillation, the actuation point – the point $u(t)$ – is motionless. These elasticities are the analogue of the one-way drive elasticities in the PEA system: they generate a low-duty cycle profile in actuator velocity, $\dot{u}(t)$, rather than in the actuator load, $F(t)$. The particular connection between $F(t)$ and $\dot{u}(t)$ is that they are the two multiplicative factors that make up the mechanical power: $P(t) = F(t)\dot{u}(t)$. The dwell-time property of these resonant states may be of some utility. For instance, it allows prolonged contact between the moving point of actuation, $u(t)$, and another structure fixed in the external reference frame – potentially facilitating the transmission of information, energy, fuel, or material to the system each cycle. This property can be replicated in dissipation-dominated systems – except, at a state of energetic neutrality rather than optimality – and it retains the characteristic SEA advantage of invariant electrical resistive losses ($\overline{P}_{F^2}$). Overall, SEA systems show lesser variety in the types of optimal elasticity than PEA systems, but maintain invariance in a wider range of metrics.

## 4. Implications and discussion

### 4.1 Implications for optimal design of biomimetic propulsion systems

The principles of elastic optimality – *i.e.*, the elastic-bound conditions – derived in Sections 2.2-2.3 have significant implications for the design of energetically-optimal forced-oscillation systems: among them, many forms of biomimetic propulsion system, such as FW-MAVs. In Section 3 we detailed some of these implications: we studied a range of particular forms of elasticity, and showed how they could be used to beneficially alter a wide range of system properties. We demonstrated a simultaneous optimisation of the system quasistatic response; control of system equilibria stability and location; and a reduction in actuator duty cycle – all while maintaining a state of minimum actuator mechanical power consumption. We also showed, in Section 2.4, how a range of additional power consumption metrics – including electrical resistive power losses – could be kept invariant under the choice of elasticity; though prioritising this invariance can sometimes limit the degree of other system alteration that is available. Nevertheless, under either prioritisation, improvements in system performance are available. These principles of elastic optimality are of great utility as design tools: they allow wide-ranging control of the system behaviour, while not compromising energetic optimality.

Indeed, to the best of our knowledge, some of the system properties generated via these principles of optimality are radically novel – for instance, the properties of the one-way drive systems (Section 3.2). These systems allow a single unidirectional actuator to generate energetically-optimal resonant oscillations, using an actuation duty cycle of 50% or lower. They have extensive application: we highlight, in particular, the possibility of designing an FW-MAV system using only a unidirectional linear actuator – *e.g.*, a single solenoid, pneumatic cylinder, or micro-combustion cylinder [84] – to generate energetically-optimal resonant oscillations. This possibility has cascading implications. For instance, unidirectional actuators of this form may show improved fuel-weight ratio relative to the current paradigm of battery-operated piezoelectric or DC motor actuation [85]. While reliable micro-combustion cylinders are not yet light enough for FW-MAV applications, the reduction in requirement to only a single cylinder (while still maintaining a form of energetic optimality) reduces the barriers for their utilisation. In this way, the one-way drive systems – and the elastic-bound principles more broadly – provide avenues for the design of radically-novel biomimetic propulsion systems.

## 4.2 Implications for solving absolute-work minimisation problems

The work-loop analysis approach that we have presented in this paper is a powerful and general tool for analysis the optimality of actuated, oscillating, systems with respect to mechanical power. Even in a highly general PEA or SEA system, the effect of nonlinear elasticity on mechanical power consumption can be decoupled, and the exact conditions for optimality can be defined. Problems of this nature, involving the minimisation of absolute work, or power – our power metric (**b**) – can sometimes be difficult to solve: the non-smoothness of the absolute-value function ($|\cdot|$) can impede numerical optimisation methods [39, 42, 43, 86]. The analysis approach that we present is notable in that solutions for states of optimal mechanical power consumption were (**i**) accessible analytically, and (**ii**) simultaneously available for a wide range of different non-smooth metrics based on mechanical power (Section 2.1). Our analysis approach was able to bypass problems of smoothness via the principle that, if the system net power is invariant with respect to a given parameter (*e.g.*, structural elasticity), then an energetically-optimal state (*e.g.*, in absolute power) is described by the condition that the system power requirement should contain no negative power (*i.e.*, global resonance).

This principle is generalisable. For instance, consider some highly general 1DOF system with a prescribed kinematic output $x(t)$, power requirement $P(t)$, and two power metrics: net power, $\overline{P}$, and absolute power, $\overline{P}_{|\cdot|}$, both defined over some $t \in [0, T]$:

$$x(t) = \tilde{x}(\lambda, t), \quad P(t) = \tilde{P}(\lambda, x(t)),$$
$$\overline{P} = \int_0^T P(t)\, dt, \quad \overline{P}_{|\cdot|} = \int_0^T |P(t)|\, dt. \tag{33}$$

$\lambda \in \mathbb{R}^N$ is a vector of $N$ time-invariant control parameters, influencing (in general) both the kinematic output $x(t)$ and the power requirement $P(t)$, via continuous real-valued functions $\tilde{x}(\cdot)$ and $\tilde{P}(\cdot)$. It is a simple property of the absolute value that $\overline{P}_{|\cdot|} \geq \overline{P}$ under any conditions. In the analysis set out in this paper, we had $\lambda$ as an infinite-dimensional parameter space: the space of all possible continuous elastic load functions, $F_s(x)$, and by the structure of $\tilde{P}(\cdot)$, we could determine that $\overline{P}$ was independent of $F_s(x)$. Now, in the general case, $\overline{P}$ is not independent of $\lambda$. But, assume we can identify a region, $\Omega$, in the control configuration space, $\Omega \subseteq \mathbb{R}^N$, such that over $\lambda \in \Omega$, $\overline{P}$, is constant. That is, $\Omega$ is a level set in metric $\overline{P}$. Then, over $\lambda \in \Omega$, if there exist any $\lambda = \lambda^*$ such that $\overline{P}_{|\cdot|} = \overline{P}$, then, at these $\lambda^*$, $\overline{P}_{|\cdot|}$ is minimised over $\Omega$. That is, these control configurations, $\lambda^*$, ensure energetic optimality in $\overline{P}_{|\cdot|}$ over $\Omega$. In addition, the condition $\overline{P}_{|\cdot|} = \overline{P}$ is equivalent to $P(t) \geq 0$, $\forall t \in [0, T]$ (*i.e.*, no negative power): this latter condition also defines $\lambda^*$,

This formalised principle is just an abstract statement of the logic used in Sections 2.2-2.3. As regards its practical application: consider, for instance, $\lambda$ as set of parameters describing the gait of a legged robot. Then, following this principle, if we can construct a level set of gait parameters, $\Omega$, such that $\lambda \in \Omega$ represents a state of constant net power, $\overline{P}$, then, if we can find the state of no negative work, $\lambda : P(t) \geq 0, \forall t$, within $\Omega$, then this state represents an energetically-optimal gait over $\Omega$. This provides an alternative, unstudied, approach to gait optimisation – and an approach which may generalise to other absolute-work minimisation problems.

**4.3 Implications for the role of structural elasticity in biological propulsion systems**

It is reasonably common, as a first-order approximation, to analyse the energetic behaviour of biological propulsion systems in terms of linear elasticity, and linear resonance – for instance, as has been done for several insect species [15, 22, 87]. There is, however, increasing evidence for nonlinear elasticities, *e.g.*, within insect flight motors [12, 13]. The energetic effects of this nonlinearity are not yet clear – but the broad theoretical analysis that we have presented sheds some light on the space of possibility. We have demonstrated that such nonlinear elasticity can be entirely consistent with energetic optimality, even in a system which is otherwise completely linear. The postulate that a biological propulsion system must be a state of energetic optimality – a broadly reasonable postulate – does not necessarily indicate that it is at a state of linear resonance, even if the system is modelled as a completely linear process.

But the implications of our theoretical results go further. In Section 3 we demonstrated that the system elasticity could beneficially alter a wide range of system properties – *e.g.*, reducing the actuator duty cycle, changing system equilibria, and even generating unidirectional load requirements – while still maintaining energetic optimality. Indeed, we found that we could alter these other system properties, in an energetically neutral way, in a system which was completely dominated by dissipative forces – a system for which no resonant energy savings are possible. The role of structural elasticity in biological propulsion systems may be as much the alteration of these system properties (*e.g.*, the synchronisation of load requirements with muscular capability), as it is any form of resonant energy saving. Our theoretical results provide a new avenue for characterising the role of elasticity in such systems.

**4.4 Future directions**

In this work, we studied systems that were (in general) strongly nonlinear, but showing a relatively simple architecture: 1DOF conventional PEA or SEA systems, with several restrictions on the work-loop behaviour of the system under analysis (Sections 2.2-2.3). The limitations of this current analysis point to future directions for theoretical development: extensions to multi-degree-of-freedom systems; systems with work loops that are non-simple curves; SEA systems with non-invertible elastic profiles; and PEA-SEA coupled systems. In addition, there is scope for the application of these theoretical techniques to a range of different systems, and for the solution of several practical problems. In particular, practical utilisation of the current results requires a precise knowledge of the system work loop; and that is predicated on a precise knowledge of the system damping – which may be unavailable. Devising control architectures for locating and controlling these global-resonant states is a crucial topic for further research, and may be a condition for the utility of these results in several systems typically showing high uncertainty (*e.g.*, FW-MAVs). Linear and nonlinear optimal control theory is obvious direction towards constructing these control architectures; but an intriguing additional place to look is within insect flight motors: elucidating the energetic role of known nonlinear elasticities in the insect flight motor might both inspire biomimetic forms of energetically-optimal biomimetic propulsion system, and shed light into the mechanical principles underlying insect flight.

## 5. Conclusions

In this work, we have presented a simple, general, and intuitive principle for constructing energetically-optimal elasticities, in a range of general dynamical systems. This principle is based on elastic bounds within a system work loop, and has diverse implications. It applies to a wide range of systems, linear and nonlinear, with parallel-elastic or series-elastic actuation.

It extends, in many ways, further into the energetic behaviour of the system, and can be used to optimise more complex metrics of power consumption (*e.g.*, metabolic, or electrical metrics). It provides a new method for designing energetically-optimal nonlinear systems; allowing design control of a range of system properties – actuator duty cycle, system equilibrium location, quasistatic behaviour, and more; all while resonant energy savings are maintained. And, finally, it provides avenues for the design of radically novel biomimetic propulsion systems: energetically-optimal systems actuated by unidirectional linear actuators, such as single solenoids or micro-combustion cylinders. Overall, this work-loop approach, and associated principles, are fundamental contributions to the study of energetic optimality in forced dynamical systems. They offer several promising avenues for further development – including generalisation and extension of the fundamental principle; and practical application in a wide range of systems and contexts.

**Declarations:**

**Acknowledgements**   This work was supported by the Azrieli Foundation Faculty Fellowship and by the Israel Ministry of Science and Technology.

**Data availability**   No datasets were generated or utilised in this study.

**Conflict of interest**   The authors declare that they have no conflict of interest.

# Appendix

## A.1 Example constructions in PEA systems

For illustration, we derive specific forms of $G^\pm(x)$, and the associated elastic optimality condition, for two different PEA systems. In the first instance, consider a linear, 1DOF, PEA, system, with time variable $t$, displacement $x$, inertia $m$, and dissipation coefficient $c$ – as per Figure 2 in the main text. The system is driven by an actuator, generating simple harmonic oscillation:

$$x(t) = \hat{x}\cos(\omega t), \quad \dot{x}(t) = -\hat{x}\omega\sin(\omega t), \quad \ddot{x}(t) = -\hat{x}\omega^2\cos(\omega t), \tag{A.1.1}$$

The required actuator load for this oscillation, inelastic ($G$) and elastic ($F$), is given by:

$$G(t) = m\ddot{x}(t) + c\dot{x}(t),$$
$$F(t) = m\ddot{x}(t) + c\dot{x}(t) + F_s(x(t)), \tag{A.1.2}$$

thus:

$$G(t) = -m\hat{x}\omega^2\cos(\omega t) - c\hat{x}\omega\sin(\omega t),$$
$$F(t) = -m\hat{x}\omega^2\cos(\omega t) - c\hat{x}\omega\sin(\omega t) + F_s(x(t)). \tag{A.1.3}$$

Parameterising $G(t)$ and $F(t)$ in terms of $x(t) = \hat{x}\cos(\omega t)$, and splitting into two solutions (arcs) we have:

$$G^\pm(x) = -m\omega^2 x \pm c\omega\sqrt{\hat{x}^2 - x^2}.$$
$$F^\pm(x) = F_s(x) - m\omega^2 x \pm c\omega\sqrt{\hat{x}^2 - x^2}. \tag{A.1.4}$$

The optimality principle then reads: for $x \in [-\hat{x}, \hat{x}]$,

$$-m\omega^2 x - c\omega\sqrt{\hat{x}^2 - x^2} \leq -F_s(x) \leq -m\omega^2 x + c\omega\sqrt{\hat{x}^2 - x^2}. \tag{A.1.5}$$

The optimal linear, cubic, and quintic elasticities – if they exist, which is dependent on the system parameters – are thus:

$$F_{s,1}(x) = m\omega^2 x,$$
$$F_{s,3}(x) = m\omega^2 x^3/\hat{x}^2, \quad\quad\quad\quad\quad\quad\quad\quad\quad (A.1.6)$$
$$F_{s,5}(x) = m\omega^2 x^5/\hat{x}^4.$$

Or, with more generality, we may define two families of polynomial elasticities, interpolated between linear and cubic, and linear and quintic profiles, respectively:

$$F_{s,1,3}(x,\alpha) = (1-\alpha)m\omega^2 x + \alpha m\omega^2 x^3/\hat{x}^2, \quad \alpha \in [0,1],$$
$$F_{s,1,5}(x,\beta) = (1-\beta)m\omega^2 x + \beta m\omega^2 x^5/\hat{x}^4, \quad \beta \in [0,1]. \quad (A.1.7)$$

The states $\alpha = 0$ and $\beta = 0$ represent linear profiles; and $\alpha = 1$ and $\beta = 1$, purely cubic and quintic profiles, respectively. The interpolated profiles, $\alpha \in [0,1]$ and $\beta \in [0,1]$, will satisfy the optimality condition, Eq. A.5, up to certain limit values of $\alpha$ and $\beta$ (possibly $\alpha = 1$ or $\beta = 1$); values which can be observed visually from plots on the work loop, or computed numerically.

In the second instance, consider a nonlinear, 1DOF, PEA system, with quadratic damping – a simplified model of an insect wing:

$$G(t) = m\ddot{x}(t) + \text{sign}(\dot{x}(t))\, c\dot{x}(t)^2,$$
$$F(t) = m\ddot{x}(t) + \text{sign}(\dot{x}(t))\, c\dot{x}(t)^2 + F_s(x(t)), \quad\quad (A.1.8)$$

thus:

$$G(t) = -m\hat{x}\omega^2 \cos(\omega t) - \text{sign}(\sin(\omega t))\, c\hat{x}^2\omega^2 \sin^2(\omega t),$$
$$F(t) = -m\hat{x}\omega^2 \cos(\omega t) - \text{sign}(\sin(\omega t))\, c\hat{x}^2\omega^2 \sin^2(\omega t) + F_s(x(t)). \quad (A.1.9)$$

Parameterising $G(t)$ and $F(t)$ in terms of $x(t) = \hat{x}\cos(\omega t)$, and splitting into two solutions (arcs) we have:

$$G^\pm(x) = -m\omega^2 x \pm c\omega^2(x^2 - \hat{x}^2),$$
$$F^\pm(x) = -m\omega^2 x \pm c\omega^2(x^2 - \hat{x}^2) + F_s(x(t)). \quad\quad (A.1.10)$$

The elastic-bound condition then reads: for $x \in [-\hat{x}, \hat{x}]$,

$$-m\omega^2 x - c\omega^2(x^2 - \hat{x}^2) \leq -F_s(x) \leq -m\omega^2 x + c\omega^2(x^2 - \hat{x}^2). \quad (A.1.11)$$

Note that, because linear inertia and harmonic motion is common to both this nonlinear system, and the linear system of Eq. A.2, the profiles Eq. A.6 are same optimal linear, cubic, and quintic profiles (if any exist). Further, the interpolated profiles of Eq. A.7 also necessarily satisfy the optimality principle, Eq. A.10, up to certain values of $\alpha$ and $\beta$, values which will differ from the linear case due to the presence of nonlinear damping.

## A.2 Example constructions in SEA systems

For illustration, we derive specific forms of $X^\pm(x)$, and the associated elastic optimality condition, for an SEA system – just one, because of the additional complexity: the process may be extended. Consider a linear, 1DOF, SEA, system, with time variable $t$, displacement $x$, inertia $m$, and dissipation coefficient $c$ – as per Figure 4 in the main text. The system is driven by an actuator, generating simple harmonic oscillation:

$$x(t) = \hat{x}\cos(\omega t), \quad \dot{x}(t) = -\hat{x}\omega\sin(\omega t), \quad \ddot{x}(t) = -\hat{x}\omega^2\cos(\omega t), \quad (A.2.1)$$

The equation of motion for this system is:

$$m\ddot{x}(t) + c\dot{x}(t) = F_s(u(t) - x(t)), \tag{A.2.2}$$

and thus, proving that $F_s(\cdot)$ is invertible, the required actuator displacement, $u(t)$, and actuator load, are given by:

$$u(t) = x(t) + F_s^{-1}(m\ddot{x}(t) + c\dot{x}(t)),$$
$$F(t) = F_s(u(t) - x(t)) = m\ddot{x}(t) + c\dot{x}(t). \tag{A.2.3}$$

It follows that the actuator displacement requirement may be alternately cast as:

$$u(t) = x(t) + F_s^{-1}(F(t)), \tag{A.2.4}$$

To reformulate this requirement entirely as $u(F)$, we must reformulate $x(t)$ in terms of $F$. From the definition of the actuator load, we have:

$$F(t) = -m\hat{x}\omega^2 \cos(\omega t) - c\hat{x}\omega \sin(\omega t), \tag{A.2.5}$$

and thus, as with the PEA, a work loop representation, $F$-$x$:

$$F^{\pm}(x) = -m\omega^2 x \pm c\omega\sqrt{\hat{x}^2 - x^2}. \tag{A.2.6}$$

To invert this into a loop $x$-$F$, we go through the following process. Expanding the square root term in both $F^+(x)$ and $F^-(x)$ yields a single multivariable polynomial, describing the elliptical shape of the work loop, which can be solved for either $F$ or $x$:

$$(m^2\omega^4 + c^2\omega^2)x^2 + 2Fm\omega^2 x + (F^2 - c^2\omega^2\hat{x}^2) = 0. \tag{A.2.7}$$

Note that $F$ has replaced $F^{\pm}$ now that these branches are unified. The solution to this polynomial, in terms of $x$, are the functions we denote $X^{\pm}(F)$:

$$X^{\pm}(F) = \frac{-mF \pm \frac{c}{\omega}\sqrt{\hat{x}^2\omega^2(m^2\omega^2 + c^2) - F^2}}{m^2\omega^2 + c^2}. \tag{A.2.8}$$

The appropriate $F$-range for $X^{\pm}(F)$, $[-\hat{F}, \hat{F}]$, is given by the extrema of Eq. A.2.6:

$$\hat{F} = \omega\hat{x}\sqrt{m^2\omega^2 + c^2}. \tag{A.2.9}$$

The derivatives $X'^{\pm}(F)$ can be computed directly from Eq. A.2.8, as:

$$X'^{\pm}(F) = \frac{1}{m^2\omega^2 + c^2}\left(-m \mp \frac{cF}{\omega\sqrt{\hat{x}^2\omega^2(m^2\omega^2 + c^2) - F^2}}\right), \tag{A.2.10}$$

and the elastic bound-conditions read:

$$X'^{-}(F) \leq -(F_s^{-1})'(F) \leq X'^{+}(F), \quad \forall F \in [F_1, 0],$$
$$X'^{+}(F) \leq -(F_s^{-1})'(F) \leq X'^{-}(F), \quad \forall F \in [0, F_2]. \tag{A.2.11}$$

We note that this state condition can be fulfilled by some $F_s^{-1}(F)$: the global-resonant state is accessible. More specifically:

$$X'^{-}(0) = X'^{+}(0) = \frac{-m}{m^2\omega^2 + c^2} \therefore (F_{s,1}^{-1})'(0) = \frac{m}{m^2\omega^2 + c^2}, \tag{A.2.11}$$

where $F_{s,1}(\cdot)$ is the system optimal linear elasticity. This value is the optimal linear compliance – leading to an optimal linear elasticity of the form:

$$F_{s,1}(u - x) = \frac{m^2\omega^2 + c^2}{m}(u - x). \tag{A.2.11}$$

Finally, for comparison, consider the case of a canonical oscillator, with an optimal linear spring. That is, we have $F_s(\delta) = k_1\delta$ (Eq. A.25); inertia and displacement are normalised ($m = 1$, $\hat{x} = 1$); and we define two canonical parameters: the undamped natural frequency, $\omega_0^2 = k_1$, and the damping ratio, $\zeta$ such that $c = 2\zeta\omega_0$. This yields the familiar EOM:

$$\ddot{x} + 2\zeta\omega_0\dot{x} + \omega_0^2 x = \omega_0^2 u. \tag{A.2.12}$$

A relationship can be derived between (**i**) the undamped natural frequency, $\omega_0$, and (**ii**) the oscillatory frequency for which $k_1$ minimises mechanical power consumption; that is, $\omega$, the initially-prescribed frequency. This relationship is:

$$\omega_0^2 = k_1 = 1/\left(F_{s,1}^{-1}\right)'(0) \therefore \omega = \omega_0\sqrt{1 - 4\zeta^2}. \tag{A.2.13}$$

This frequency is the global-resonant frequency (*i.e.*, energetically optimal) frequency for the SEA system: it differs slightly from both the damped natural frequency (root factor $1 - \zeta^2$) and the linear resonant frequency (root factor $1 - 2\zeta^2$).

### A.3 Proof of mechanical power optimality in PEA systems

In this section, we prove the PEA elastic optimality results that we have utilised in the main text, and prior to this point in the appendix. We begin at the formulation of the system as per the work loop of Eq. 7:

$$\begin{aligned} G^\pm(x) &= G_{\text{mid}}(x) \pm G_{\text{arc}}(x), \\ F^\pm(x) &= G_{\text{mid}}(x) \pm G_{\text{arc}}(x) + F_s(x), \end{aligned} \tag{A.3.1}$$

over $x \in [x_1, x_2]$. If a system can be formulated as a work loop of this form, respecting the condition that the loop must be a simple closed curve, no more than bivalued at any $x$, and showing net power dissipation, then the system is admissible under this analysis.

First, consider evaluating the four integrals of mechanical power, Eq. 8, for $F^\pm(x)$. For the net power, $\overline{P}_{(a)}$, we have the result:

$$\overline{P}_{(a)} = \frac{1}{T}\int_{x_1}^{x_2} F^+(x) - F^-(x)\, dx = \frac{2}{T}\int_{x_1}^{x_2} G_{\text{arc}}(x)\, dx; \tag{A.3.2}$$

that is, the net power – the area within the work loop – is unaffected by elasticity, $F_s(x)$, exactly as we would expect. For the other metrics, $\overline{P}_{(b)}$, $\overline{P}_{(c)}$ and $\overline{P}_{(d)}$, however, no cancellation or simplification occurs, and a general dependency on elasticity, $F_s(x)$, remains. To gain further insight into the behaviour of these latter metrics with respect to $F_s(x)$, consider an alternative parameterisation of the work loop. Split $F^+(x)$ and $F^-(x)$ into four components, representing regions of load above ($A$) and below ($B$) zero, in the following way.

$$\begin{aligned} F^+(x) &= A^+(x) - B^+(x), & F^-(x) &= A^-(x) - B^-(x), \\ A^+(x) &= F^+(x)[F^+(x) \geq 0]_{\mathbb{I}}, & A^-(x) &= F^-(x)[F^-(x) \geq 0]_{\mathbb{I}}, \\ B^+(x) &= -F^+(x)[F^+(x) \leq 0]_{\mathbb{I}}, & B^-(x) &= -F^-(x)[F^-(x) \leq 0]_{\mathbb{I}}. \end{aligned} \tag{A.3.4}$$

Thus, $A^{\pm}(x), B^{\pm}(x) \geq 0, \forall x$, and individual power integrals for each component are necessarily positive:

$$W(I) = \frac{1}{T}\int_{x_1}^{x_2} I(x)\,dx \geq 0, \qquad I(x) \in \{A^+(x), B^+(x), A^-(x), B^-(x)\}. \tag{A.3.5}$$

This allows evaluation of the four power metrics, Eq. 8, in terms of these $W(I)$. For the net power, $\overline{P}_{(a)}$, we have the result:

$$\overline{P}_{(a)} = W(A^+) + W(B^-) - W(A^-) - W(B^+), \tag{A.3.6}$$

where the terms $W(A^-)$ and $W(B^+)$ serve to reduce the actuator power consumption – this represents the storage and release of negative power. In addition, we have explicit results for $\overline{P}_{(b)}$, $\overline{P}_{(c)}$ and $\overline{P}_{(d)}$:

$$\begin{aligned}\overline{P}_{(b)} &= W(A^+) + W(B^-) + W(A^-) + W(B^+),\\ \overline{P}_{(c)} &= W(A^+) + W(B^-),\\ \overline{P}_{(d)} &= W(A^+) + W(B^-) - W(A^-) - W(B^+) + W_Q,\end{aligned} \tag{A.3.7}$$

with the special penalty integral for $\overline{P}_{(d)}$:

$$W_Q = \frac{1}{T}\int_{x_1}^{x_2} Q^+ F^+ [F^+ \leq 0]_{\mathbb{I}}\,dx + \frac{1}{T}\int_{x_1}^{x_2} Q^- F^- [F^- \geq 0]_{\mathbb{I}}\,dx. \tag{A.3.8}$$

These explicit forms have several implications, representable in the form of the following inequalities, valid $\forall F^{\pm}(x)$:

$$\begin{aligned}\overline{P}_{(b)} &\geq \overline{P}_{(a)},\\ \overline{P}_{(c)} &\geq \overline{P}_{(a)},\\ \overline{P}_{(b)} &\geq \overline{P}_{(c)},\\ \overline{P}_{(d)} &\geq \overline{P}_{(a)},\end{aligned} \tag{A.3.9}$$

these results can be seen from the fact that $W(I) > 0, \forall I(x)$, and in the case of $\overline{P}_{(d)} \geq \overline{P}_{(a)}$, from the fact that $Q^{\pm}(x) > 0$ necessitates $W_Q \geq 0$.

Synthesizing the information from Eqs. A.3.2 and A.3.9 yields the following conclusions. From Eq. A.3.2, $\overline{P}_{(a)}$ is independent of $F_s(x)$; whereas $\overline{P}_{(b)}$, $\overline{P}_{(c)}$ and $\overline{P}_{(d)}$, are (in general) dependent. From Eq. A.3.9, the minimum possible value that $\overline{P}_{(b)}$, $\overline{P}_{(c)}$ and $\overline{P}_{(d)}$ can take, under any $F_s(x)$, is this independent value of $\overline{P}_{(a)}$. There is no immediate guarantee that this minimum is attainable; but we may explore the conditions for it to be attained:

$$\begin{aligned}\overline{P}_{(b)} &= \overline{P}_{(a)} \therefore W(A^-) + W(B^+) = 0 \therefore W(A^-) = W(B^+) = 0,\\ \overline{P}_{(c)} &= \overline{P}_{(a)} \therefore W(A^-) + W(B^+) = 0 \therefore W(A^-) = W(B^+) = 0,\\ \overline{P}_{(d)} &= \overline{P}_{(a)} \therefore W_Q = 0.\end{aligned} \tag{A.3.10}$$

Consider these two sets of conditions separately. For $\overline{P}_{(b)}$ and $\overline{P}_{(c)}$, given that $A^-(x)$ and $B^+(x)$ cannot change sign (Eq. A.3.4), the integral condition $W(A^-) = W(B^+) = 0$ implies the functional condition $A^-(x) = B^+(x) = 0, \forall x$; that is,, $\forall x$:

$$\begin{aligned}F^+(x)[F^+(x) \leq 0]_{\mathbb{I}} &= 0,\\ F^-(x)[F^-(x) \geq 0]_{\mathbb{I}} &= 0.\end{aligned} \tag{A.3.11}$$

For $\overline{P}_{(d)}$, the condition $W_Q = 0$ implies directly the identical condition, that, $\forall x$:

$$F^+(x)[F^+(x) \leq 0]_\mathbb{I} = 0,$$
$$F^-(x)[F^-(x) \geq 0]_\mathbb{I} = 0. \tag{A.3.12}$$

It follows that this single pair of conditions, Eqs. A.3.11 or A.3.12, ensure optimality in mechanical power both necessarily and sufficiently: they are the sole conditions to ensure that the inequalities of Eq. A.3.9 become equalities, and thus the mechanical power metrics take their minimum possible values.

These conditions are interpretable in terms of $F_s(x)$, in the following way. To ensure that the expressions in Eqs. A.3.11 or A.3.12 are zero-valued $\forall x$, the following must be true:

$$\forall x \text{ either } F^+(x) = 0 \text{ or } [F^+(x) \leq 0]_\mathbb{I} = 0,$$
$$\forall x \text{ either } F^-(x) = 0 \text{ or } [F^-(x) \geq 0]_\mathbb{I} = 0. \tag{A.3.13}$$

Computing the Iverson bracket, these conditions become:

$$\forall x \text{ either } F^+(x) = 0 \text{ or } F^+(x) > 0,$$
$$\forall x \text{ either } F^-(x) = 0 \text{ or } F^-(x) < 0, \tag{A.3.14}$$

or, as two simple inequalities:

$$F^+(x) \geq 0,$$
$$F^-(x) \leq 0. \tag{A.3.15}$$

This is the intuitive result of Section 2.2: the presence of negative power in the system necessarily implies that the system is suboptimal in terms of mechanical power, or mechanical efficiency. An elasticity that ensures that the resultant actuator work loop contains no region of negative power will be optimal.

What are the conditions on $F_s(x)$ to ensure Eq. A.3.15 is satisfied? This is simple to compute. As $F^\pm(x) = G^\pm(x) + F_s(x)$, by the system definition, Eq. A.3.1, it follows that:

$$G^+(x) + F_s(x) \geq 0,$$
$$G^-(x) + F_s(x) \leq 0, \tag{A.3.16}$$

and thus:

$$G^-(x) \leq -F_s(x) \leq G^+(x), \tag{A.3.17}$$

which is the elastic-bound condition, Eq. 10, which we stated in Section 2.2. In this proof, we have demonstrated that this condition is necessary and sufficient to ensure that any of the mechanical power metrics, $\overline{P}_{(b)}$, $\overline{P}_{(c)}$ and $\overline{P}_{(d)}$, are minimised. The sufficiency of this condition, at least, can be verified by numerical tests. Note, however, that (**i**) we have computed this optimality condition analytically, not numerically, in contrast to most existing treatments of absolute-power minimization in the literature [39, 40, 45]; (**ii**) we required no calculus of optimization to do so: optimality can be deduced simply from the form of the power integrals in the $F$-$x$ domain; and (**iii**) no differentiability requirements have been imposed on any system parameters. Analysis in the $F$-$x$ domain is a powerful tool for analysing the optimality of system parameters with respect to mechanical power consumption, and these techniques may be applicable to problems of absolute-power minimisation that have previously been treated numerically.

## A.4 Proof of mechanical power optimality in SEA systems

We begin with a general SEA system defined as per Eq. 12, with no other restrictions yet placed on it. We then place the following restrictions, defining the admissibility of the system for our analysis. We require that there exist time-domain system functions $x(t)$, $\dot{x}(t)$, $F(t)$, and $\dot{F}(t)$, all continuous and real-valued over all real-valued $t$, and periodic with period $T$. Consider the function $\dot{F}(t)$ in more detail. We define two sets of times, $T^+$ and $T^-$, in relation to this function: $T^+ = \{t : \dot{F}(t) < 0\}$, and $T^- = \{t : \dot{F}(t) \geq 0\}$. Note the allocation of '+' with '< 0' and '−' with '≥ 0' is intentional. We require that these sets of times exist. If they do, they are necessarily unique, and together span all $t$, i.e., $T^+ \cup T^- = \mathbb{R}$. We can then use these time windows to segment $\dot{F}(t)$ into two functions: $\dot{f}^+(t) = \dot{F}(t)$, defined for $t \in T^+$, and $\dot{f}^-(t) = \dot{F}(t)$, defined for $t \in T^-$. It follows that $\dot{f}^+(t) < 0, \forall t \in T^+$, and $\dot{f}^+(t) \geq 0, \forall t \in T^+$, and we may reconstruct $\dot{F}(t)$ as:

$$\dot{F}(t) = \begin{cases} \dot{f}^+(t) & t \in T^+, \\ \dot{f}^-(t) & t \in T^-. \end{cases} \tag{A.4.1}$$

We now apply the same process to $x(t)$, $\dot{x}(t)$ and $F(t)$:

$$x(t) = \begin{cases} x^+(t) & t \in T^+, \\ x^-(t) & t \in T^-, \end{cases} \quad \dot{x}(t) = \begin{cases} \dot{x}^+(t) & t \in T^+, \\ \dot{x}^-(t) & t \in T^-, \end{cases}$$

$$F(t) = \begin{cases} f^+(t) & t \in T^+, \\ f^-(t) & t \in T^-, \end{cases} \tag{A.4.2}$$

defining $x^{\pm}(t)$, $\dot{x}^{\pm}(t)$ and $F^{\pm}(t)$. We require the following condition be satisfied by these functions: that the value of $x^+(t)$, $x^-(t)$, $\dot{x}^+(t)$ and $\dot{x}^-(t)$, at all $t$, must be uniquely defined by the value $F(t)$. In equivalent terms: $f^+(t)$ and $f^-(t)$ must be invertible, i.e., the periodic waveform $F(t)$ must be composed to two monotonic half cycles.

If this condition is satisfied, we may parameterise $x^{\pm}(t)$ and $\dot{x}^{\pm}(t)$ in terms of $F$:

$$x(t) = \begin{cases} X^+(F(t)) & t \in T^+, \\ X^-(F(t)) & t \in T^-, \end{cases} \quad \dot{x}(t) = \begin{cases} \dot{X}^+(F(t)) & t \in T^+, \\ \dot{X}^-(F(t)) & t \in T^-, \end{cases}$$

$$\dot{F}(t) = \begin{cases} \dot{F}^+(F(t)) & t \in T^+, \\ \dot{F}^-(F(t)) & t \in T^-, \end{cases} \tag{A.4.3}$$

where $x^{\pm}(t)$ and $\dot{x}^{\pm}(t)$ have now been reformulated into the functions $X^{\pm}(F)$, $\dot{X}^{\pm}(F)$ and $\dot{F}^{\pm}(F)$, defined over an interval in force, $F \in [F_1, F_2]$, where $F_1 = \min F(t)$ and $F_2 = \max F(t)$. The function $X^{\pm}(F)$ we will recognise as a work loop: $dx \cdot dF$ is the differential of work. The final condition which we require to be satisfied is: $X^+(F) \geq X^-(F), \forall F \in [F_1, F_2]$ (NB: the motivation behind the allocation of the superscript symbols, $\pm$). We note that the conditions $X^+(F_1) = X^-(F_1)$ and $X^+(F_2) = X^-(F_2)$ are already necessarily satisfied by nature of the continuity of $x(t)$.

If all these conditions are satisfied, then we consider the system admissible for our analysis, and we may proceed. Consider an elastic element in the system, with elastic load profile $F_s(\delta)$, continuous, differentiable, and real-valued over all real-valued $\delta$. We require that $F_s(\delta)$ be invertible. We may then express the SEA dynamics and power requirements (cf. Eq. 13) as:

$$u(t) = x(t) + F_s^{-1}(F(t)),$$
$$\dot{u}(t) = \dot{x}(t) + \dot{F}(t)(F_s^{-1})'(F(t)), \tag{A.4.4}$$
$$P(t) = F(t)\dot{u}(t),$$

where $(F_s^{-1})'(F)$ denotes $\frac{d}{dF}F_s^{-1}(F)$. Note that we have used the differential relation $dF = \dot{F}dt$ to reformulate $\frac{d}{dt}F_s^{-1}(F)$ and $\dot{F}(t)\frac{d}{dF}F_s^{-1}(F)$. Using the sets of times, $T^+$ and $T^-$, we may segment these additional functions in the manner of Eq. A.4.2:

$$u(t) = \begin{cases} u^+(t) & t \in T^+, \\ u^-(t) & t \in T^-, \end{cases} \quad \dot{u}(t) = \begin{cases} \dot{u}^+(t) & t \in T^+, \\ \dot{u}^-(t) & t \in T^-, \end{cases}$$
$$P(t) = \begin{cases} p^+(t) & t \in T^+, \\ p^-(t) & t \in T^-, \end{cases} \tag{A.4.5}$$

and parameterise them in terms of $F$, in the manner of Eq. A.4.3:

$$u(t) = \begin{cases} U^+(F(t)) & t \in T^+, \\ U^-(F(t)) & t \in T^-, \end{cases} \quad \dot{u}(t) = \begin{cases} \dot{U}^+(F(t)) & t \in T^+, \\ \dot{U}^-(F(t)) & t \in T^-, \end{cases}$$
$$P(t) = \begin{cases} P^+(F(t)) & t \in T^+, \\ P^-(F(t)) & t \in T^-, \end{cases} \tag{A.4.6}$$

Now, we can relate these segmented and parameterised functions to known system functions, Eqs. A.4.2 and A.4.3. Over $F \in [F_1, F_2]$:

$$u^\pm(t) = x^\pm(t) + F_s^{-1}(f^\pm(t)),$$
$$\dot{u}^\pm(t) = \dot{x}^\pm(t) + \dot{f}^\pm(t) \cdot (F_s^{-1})'(f^\pm(t)),$$
$$p^\pm(t) = f^\pm(t) \cdot \left(\dot{x}^\pm(t) + \dot{f}^\pm(t)(F_s^{-1})'(f^\pm(t))\right), \tag{A.4.7}$$
$$U^\pm(F) = X^\pm(F) + F_s^{-1}(F),$$
$$\dot{U}^\pm(F) = \dot{X}^\pm(F) + \dot{F}^\pm(F) \cdot (F_s^{-1})'(F),$$
$$P^\pm(F) = F \cdot \left(\dot{X}^\pm(F) + \dot{F}^\pm(F) \cdot (F_s^{-1})'(F)\right).$$

In addition, we can define the work-loop tangent functions, $X'^\pm(F) = \frac{d}{dF}X^\pm(F)$ and $U'^\pm(F) = \frac{d}{dF}X^\pm(F)$, where:

$$X'^\pm(F) = \frac{\dot{X}^+(F)}{\dot{F}^+(F)}, \quad U'^\pm(F) = X'^\pm(F) + (F_s^{-1})'(F), \tag{A.4.8}$$

and thus:

$$P^\pm(F) = F \cdot \dot{F}^\pm(F) \cdot \left(X'^\pm(F) + (F_s^{-1})'(F)\right). \tag{A.4.9}$$

We are now able to evaluate the actuator power consumption metrics **(a)**-**(d)**, as per Section 2.1, in terms of $P^\pm(F)$. For the net power, metric **(a)**:

$$\overline{P}_{(a)} = \frac{1}{T}\int_0^T P(t)\,dt = \frac{1}{T}\left(\int_{t \in T^+} p^+(t)\,dt + \int_{t \in T^-} p^-(t)\,dt\right)$$
$$= \frac{1}{T}\left(\int_{F_2}^{F_1} \frac{P^+(F)}{\dot{F}^+(F)}\,dF + \int_{F_1}^{F_2} \frac{P^-(F)}{\dot{F}^-(F)}\,dF\right), \quad \text{(A.4.10)}$$

where, flipping limits; $[F_2, F_1]$ to $[F_1, F_2]$, we obtain:

$$\overline{P}_{(a)} = \frac{1}{T}\left(\int_{F_1}^{F_2} F \cdot \left(X'^-(F) - X'^+(F)\right) dF\right). \quad \text{(A.4.11)}$$

We require that the sets of time $T^\pm$ permit an adequate definition of integration (*i.e.*, they are composed of continuous intervals). We observe that $\overline{P}_{(a)}$ is independent of $F_s(\delta)$: the net power is unaltered by elasticity. We can perform the same process for $\overline{P}_{(d)}$, and thereby recover both $\overline{P}_{(b)}$ and $\overline{P}_{(c)}$, as we already know (Section 2.1) that they are special cases of $\overline{P}_{(d)}$. Evaluating $\overline{P}_{(d)}$, we obtain:

$$\overline{P}_{(d)} = \frac{1}{T}\int_0^T (P(t) + Q(t)|P(t)|[P(t) \leq 0]_\mathbb{I})\,dt =$$
$$= \frac{1}{T}\int_{t \in T^+} (p^+(t) + q^+(t)|p^+(t)|[p^+(t) \leq 0]_\mathbb{I})\,dt \quad \text{(A.4.12)}$$
$$+ \frac{1}{T}\int_{t \in T^-} (p^-(t) + q^-(t)|p^-(t)|[p^-(t) \leq 0]_\mathbb{I})\,dt,$$

and thus $\overline{P}_{(d)} = \overline{P}_{(a)} + \Delta\overline{P}$, where:

$$\Delta\overline{P} = \frac{1}{T}\int_{F_1}^{F_2}\left(\frac{Q^-(F)|P^-(F)|[P^-(F) \leq 0]_\mathbb{I}}{\dot{F}^-(F)} - \frac{Q^+(F)|P^+(F)|[P^+(F) \leq 0]_\mathbb{I}}{\dot{F}^+(F)}\right) dt. \quad \text{(A.4.13)}$$

We can see that $\Delta\overline{P} \geq 0$ in any system that we have defined:

(**i**) For power metric (**d**), $Q(t) > 0, \forall t$, and so $Q^\pm(F) > 0, \forall F$.

(**ii**) $|P^\pm(F)| \geq 0, \forall F$, and $[P^-(F) \leq 0]_\mathbb{I} \geq 0, \forall F$, necessarily.

(**iii**) $\dot{F}^-(F) \geq 0$ and $\dot{F}^+(F) < 0$, by our construction.

Thus $\Delta\overline{P} \geq 0$. Physically, this is a simply restatement of the principle that the actuator power consumption is always greater than of equal to the net work – as studied also in appendix A.3. It follows that if we can use elasticity, $F_s(\delta)$, to ensure that $\Delta\overline{P} = 0$ (*i.e.*, $\overline{P}_{(a)} = \overline{P}_{(d)}$) then we will have reached a state of minimum $\overline{P}_{(d)}$ w.r.t. elasticity. Naturally, based on the time-domain formulation of $\overline{P}_{(a)}$ and $\overline{P}_{(d)}$, we can see that $\overline{P}_{(a)} = \overline{P}_{(d)}$ when $P(t) \geq 0, \forall t$ – *i.e.*, the absence of negative work; and the global resonance condition of Ma and Zhang [27–30]. What elasticity, $F_s(\delta)$, will ensure that this condition is satisfied? If $P(t) \geq 0, \forall t$, then $p^+(t) \geq 0, \forall t \in T^+$ and $p^-(t) \geq 0, \forall t \in T^-$; and thus, $P^\pm(F) \geq 0, \forall F \in [F_1, F_2]$. Hence:

$$F \cdot \dot{F}^{\pm}(F) \cdot \left( X'^{\pm}(F) + (F_s^{-1})'(F) \right) \geq 0, \qquad \forall F \in [F_1, F_2]. \tag{A.4.12}$$

We may disassemble this condition according to the signs of $F$ and $\dot{F}^{\pm}(F)$:

(**A**) Consider $F > 0$, i.e., $F \in (0, F_2]$. Then:

$$\dot{F}^{\pm}(F) \cdot \left( X'^{\pm}(F) + (F_s^{-1})'(F) \right) \geq 0, \qquad \forall F \in (0, F_2], \tag{A.4.13}$$

(**A.i**) Take $\dot{F}^{+}(F) > 0$, over $X'^{+}(F)$. Then:

$$\boxed{-(F_s^{-1})'(F) \geq X'^{+}(F), \qquad \forall F \in (0, F_2],} \tag{A.4.14}$$

(**A.ii**) Or, take instead $\dot{F}^{-}(F) < 0$, $\dot{F}^{-}(F) \neq 0$, over $X'^{-}(F)$. Then:

$$\boxed{-(F_s^{-1})'(F) \leq X'^{-}(F), \qquad \forall F \in (0, F_2], \qquad \dot{F}^{-}(F) \neq 0.} \tag{A.4.15}$$

(**B**) Consider the case $F < 0$, i.e., $F \in [F_1, 0)$. Then:

$$\dot{F}^{\pm}(F) \cdot \left( X'^{\pm}(F) + (F_s^{-1})'(F) \right) \leq 0, \qquad \forall F \in [F_1, 0), \tag{A.4.16}$$

(**B.i**) Take $\dot{F}^{+}(F) > 0$, over $X'^{+}(F)$. Then:

$$\boxed{-(F_s^{-1})'(F) \leq X'^{+}(F), \qquad \forall F \in (0, F_2],} \tag{A.4.17}$$

(**B.ii**) Or, take instead $\dot{F}^{-}(F) < 0$, $\dot{F}^{-}(F) \neq 0$, over $X'^{-}(F)$. Then:

$$\boxed{-(F_s^{-1})'(F) \geq X'^{-}(F), \qquad \forall F \in (0, F_2], \qquad \dot{F}^{-}(F) \neq 0.} \tag{A.4.18}$$

(**C**) Consider $F = 0$ and/or $\dot{F}^{-}(F) = 0$.

Then the elasticity is unbounded. However, note that as $F = \delta F \to 0$, from both below and above zero, then we have the four limits:

$$\begin{aligned} X'^{-}(\delta F) \leq -(F_s^{-1})'(\delta F) \leq X'^{+}(\delta F), & \qquad \delta F \to 0, \qquad \delta F < 0, \\ X'^{+}(\delta F) \leq -(F_s^{-1})'(\delta F) \leq X'^{-}(\delta F), & \qquad \delta F \to 0, \qquad \delta F > 0, \end{aligned} \tag{A.4.19}$$

and thus, if the values $X'^{\pm}(0)$ exist, then we require that $-(F_s^{-1})'(0) = X'^{\pm}(0)$. This contrasts to the behaviour of these conditions as $\dot{F} = \delta \dot{F} \to 0$: there, $X' \to \infty$ (unless, $\dot{x} \to 0$ more rapidly), and so, in general, the optimal compliance will be unbounded in this limit.

Concatenating all these conditions, we obtain the elastic-bound conditions, Eq. 20:

$$\begin{aligned} X'^{-}(F) \leq -(F_s^{-1})'(F) \leq X'^{+}(F), & \qquad \forall F \in [F_1, 0]; \\ X'^{+}(F) \leq -(F_s^{-1})'(F) \leq X'^{-}(F), & \qquad \forall F \in [0, F_2]. \end{aligned} \tag{A.4.20}$$

In this proof, we have demonstrated that these conditions are sufficient and necessary to ensure that any of the mechanical power metrics, $\overline{P}_{(b)}$, $\overline{P}_{(c)}$ and $\overline{P}_{(d)}$, take the value of the net power, $\overline{P}_{(a)}$ – a value less than or equal to their minimum possible value over the space of $F_s(\cdot)$. However, as discussed more fully in Section 2.3, it is conceivable that Eq. is impossible to satisfy, in which case the value of the net power, $\overline{P}_{(a)}$, is *less than* the minimum value(s) of $\overline{P}_{(b)}$, $\overline{P}_{(c)}$ and $\overline{P}_{(d)}$ over the space of the space of $F_s(\cdot)$. In this scenario, these minimum value(s) must be computed via some other method. The condition on $X'^{\pm}(F)$ that ensures that Eq. A.4.20 admits some solution for $(F_s^{-1})'(F)$ is simply that:

$$\begin{aligned} X'^{+}(F) \leq X'^{-}(F), & \quad \forall F \in [F_1, 0], \\ X'^{-}(F) \leq X'^{+}(F), & \quad \forall F \in [0, F_2], \end{aligned} \quad (A.4.21)$$

implying, at $F = 0$, that $X'^{+}(0) = X'^{-}(0)$. These are the system conditions discussed in Section 2.3.

## A.5 Proof of conditional absolute-load invariance in PEA systems.

The PEA elastic-bound conditions (Eq. 10) also define a region of invariance in the absolute load integral, $\overline{P}_{|F|}$, as per Section 3.2. Here we provide proof. Consider a general PEA system, as per Eq. 7, with its attendant conditions for admissibility. We may evaluate $\overline{P}_{|F|}$ by segmenting $F(t)$ into two time-domain profiles, $F^+(t)$ following $F^-(t)$, without loss of generality, representing $F^{\pm}(x)$ in the time domain. The time $T^*$ represents the transition between $F^{\pm}(t)$. Under an elasticity that is optimal in terms of mechanical power consumption, $F^+(t) \geq 0$ and $F^-(t) \leq 0$, and thus we have:

$$\overline{P}_{|F|} \propto \int_0^T |F(t)| \, dt = \int_{T^*}^T F^+(t) \, dt - \int_0^{T^*} F^-(t) \, dt. \quad (A.5.1)$$

Via the differential relation, $dt = 1/\dot{x} \, dx$, we may recast this integral into an integral over $x$, with the velocity functions $\dot{x}^+(x)$ and $\dot{x}^-(x)$ representing the velocities associated with time windows $[T^*, T]$ and $[0, T^*]$, respectively, matching $F^{\pm}(x)$. This yields:

$$\overline{P}_{|F|} \propto \int_{x_1}^{x_2} \frac{F^+(x)}{\dot{x}^+(x)} dx + \int_{x_1}^{x_2} \frac{F^-(x)}{\dot{x}^-(x)} dx, \quad (A.5.2)$$

where, note, the negative sign arises because of the need to switch integration over the window $[x_2, x_1]$ to $[x_1, x_2]$. If, then, we have a symmetric waveform, $x(t) = x(T - t)$, $\forall t$, in the time-domain, or $\dot{x}^+(x) = -\dot{x}^-(x) = \dot{x}_{\text{ref}}(x) > 0$, $x \in [x_1, x_2]$ in the displacement domain, then we have:

$$\overline{P}_{|F|} \propto \int_{x_1}^{x_2} \frac{F^+(x) - F^-(x)}{\dot{x}_{\text{ref}}(x)} dx. \quad (A.5.3)$$

Under the definition of $F^{\pm}(x)$, this reduces to:

$$\overline{P}_{|F|} \propto \int_{x_1}^{x_2} \frac{G_{\text{arc}}(x)}{\dot{x}_{\text{ref}}(x)} dx, \quad (A.5.4)$$

which is independent of $F_s(x)$. That is, under the waveform symmetry condition, $x(t) = x(T - t)$, $\forall t$, the metric $\overline{P}_{|F|}$ is invariant under any elasticity satisfying Eq. 10.

https://doi.org/10.1177/0278364908095333